\documentclass[letterpaper]{article}

\usepackage{booktabs, tabularx, array}
\usepackage[percent]{overpic}
\usepackage[T1]{fontenc}
\usepackage{geometry}
\usepackage{setspace}
\usepackage{svg}
\usepackage{subcaption}
\usepackage{graphicx}
\usepackage{placeins}
\usepackage{graphicx}
\usepackage{longtable}
\usepackage{booktabs}
\usepackage{array}
\usepackage{float}
\usepackage{placeins}
\usepackage[export]{adjustbox}
\usepackage[style=chem-acs, articletitle=true]{biblatex}
\usepackage{graphicx}
\usepackage{float}
\newfloat{scheme}{htbp}{los}
\floatname{scheme}{Scheme}
\floatname{chart}{Chart}
\newfloat{graph}{htbp}{loh}

\usepackage{chemformula} % Formulas using \ch{}
\usepackage[version = 4]{mhchem} % Formulas using \ce{}

\usepackage{cleveref}
\usepackage{authblk}
\author[1,2]{Hong-Fei LI}
\author[1,2]{Xi-Lin GAO}
\author[3]{Yi-Juan XIANG}
\author[4]{Shu-Song HUANG}
\author[5]{Yi-lin WANG}
\author[2]{Chun-Dong XUE}
\author[1]{Zhuo YANG*}
\author[1,2]{Yong-Jiang LI*}
\author[1,2]{Xu-Qu HU*}
\affil[1]{Cancer Hospital of Dalian University of Technology, Dalian University of Technology, Shenyang 110000, China}
\affil[2]{Faculty of Medicine, Dalian University of Technology, Dalian 116024, China}  %School of Biomedical Engineering, 
\affil[3]{Hunan Provincial Maternal and Child Health Care Hospital, Changsha 410000, China}
\affil[4]{CReSTIC, Université de Reims Champagne-Ardenne, Reims 51100, France}
\affil[5]{The Queen's University of Belfast Joint College, China Medical University, Shenyang 110122, China}
\title{Robust Image-Driven Phenotyping of Ovarian Tumor Cells using Optimized Dynamic Features in Hyperbolic Channels}

\date{*Email: Xu-Qu Hu (huxuqu@dlut.edu.cn); Yong-Jiang LI (yongjiangli@dlut.edu.cn); Zhuo YANG (zhuoyang@dlut.edu.cn)}
\begin{document}

\maketitle

%--------------------------------------------------------------%
%--------------------------------------------------------------%
\begin{abstract}
Image-based cellular mechanophenotyping in microfluidic devices provides a high-throughput, label-free approach for single-cell profiling. 
While complex microchannels, such as hyperbolic geometries, induce continuous extensional stress to reveal transient deformation dynamics, the resulting high-dimensional feature spaces are inherently susceptible to hydrodynamic artifacts. 
Variations in operational flow rates frequently distort discriminative boundaries, coupling feature distributions to fluidic conditions rather than intrinsic cellular biology. 
To address this limitation, we developed a stability-guided analytical framework that systematically decouples flow-induced physical noise from authentic mechanobiological signatures. 
By tracking the morphodynamic, kinematic, and intracellular optical-density trajectories of healthy and malignant ovarian cell lines, we extracted a 93-dimensional dynamic feature space. We then applied a cross-flow screening strategy—based on non-parametric structural consistency and statistical persistence—to isolate hydrodynamically robust descriptors. 
This framework successfully compressed the data into task-adapted subsets (20 features for binary cancer-versus-healthy classification; 25 features for multi-class cancer subtyping). Variance-attribution analysis confirmed that this refinement neutralized flow-conditioned spatial migration; for instance, flow-associated variance in the primary principal component decreased from 69.9\% to 9.3\% in the subtyping task. 
Compositional mapping further revealed that macroscopic binary discrimination relies primarily on bulk kinematic transitions, whereas clonal subtyping necessitates localized intracellular optical heterogeneity. 
The optimized subsets preserved diagnostic fidelity across diverse machine learning architectures and restricted data-sampling conditions. 
By rigorously penalizing flow-variant descriptors, this framework operates as a fundamental biophysical optimization protocol, establishing a reliable basis for flow-independent, continuous dynamic phenotyping.
\end{abstract}
%--------------------------------------------------------------%
%--------------------------------------------------------------%

%---------------------------------------------%
\section{Introduction}
%---------------------------------------------%

Image-based cellular phenotyping has emerged as a fundamental label-free approach for quantitative single-cell discrimination \cite{moshkov2024learning}. Unlike fluorescence-based techniques, bright-field imaging directly captures intrinsic morphological and optical-intensity features without the need for exogenous labels \cite{otto2015real, chandrasekaran2024three}. These label-free image sequences can be converted into quantitative descriptors that summarize cellular geometry, deformation, motion, and optical-intensity variation \cite{zhou2023computer}. Such descriptors are commonly integrated with machine-learning classifiers for biological profiling, primarily because their computational efficiency and physical interpretability facilitate high-throughput screening workflows \cite{zhou2023computer, chandrasekaran2024three}.

Conventional microfluidic platforms typically use uniform channel geometries (e.g., rectangular PDMS channels) to maintain steady hydrodynamic environments. This subjects cells to near-constant mechanical loading, resulting in quasi-static deformation \cite{banik2023CRBio, liu2023LabChip, zhou2024AC, smith2024MNFluids}. Within these constant cross-sections, cellular responses are evaluated via single-frame measurements or time-averaged metrics, such as cell deformation, projected area, or inferred mechanical parameters \cite{otto2015real, mietke2015extracting, mokbel2017numerical}. These quasi-static assessments inherently overlook the continuous mechanical responses of cells exposed to spatially varying hydrodynamic stresses. Spatiotemporal variations in dynamic images, including intracellular optical-intensity redistributions, contain supplementary information regarding transient cellular deformation and viscoelastic recovery \cite{chandrasekaran2024three, tegtmeyer2024high, zhou2023computer}. However, conventional quasi-static frameworks rarely incorporate these time-resolved signals.

%--------------------------------------------------%
\begin{figure}[t]
\centering{
\includegraphics[width=\linewidth]{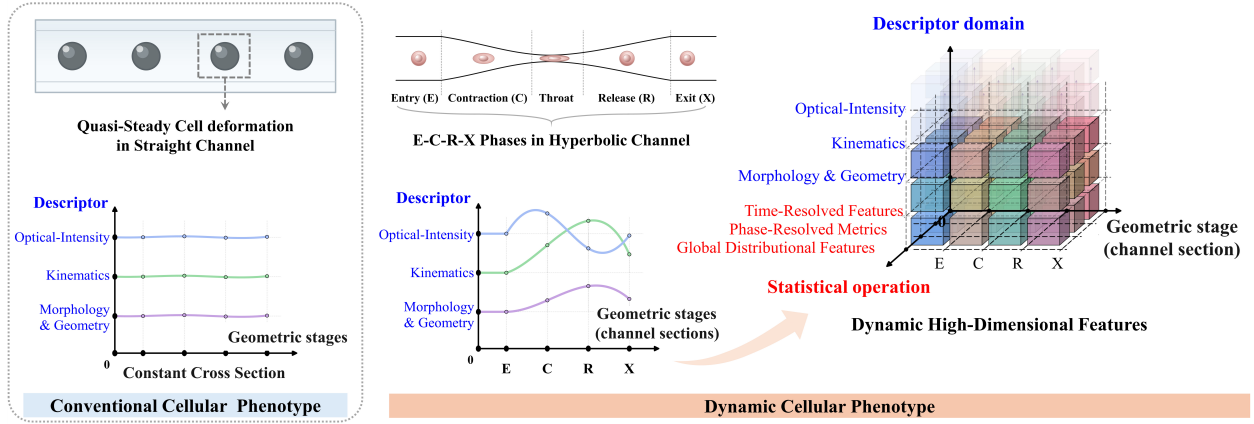}}
\caption{Schematic illustration of high-dimensional dynamic feature construction in hyperbolic contraction--expansion microchannel-based cellular phenotyping. Conventional straight channels provide a limited set of quasi-static cellular descriptors. Conversely, the varying cross-section of the hyperbolic microchannel induces continuous deformation-related responses during cell transit. Image-derived morphological, grayscale, and kinematic signals are quantified across sequential spatial stages and encoded into a structured high-dimensional dynamic feature space.}
\label{fig}
\end{figure}
%--------------------------------------------------%

To capture these transient dynamics, recent studies have shifted toward complex microchannels with varying cross-sections \cite{banik2023CRBio, liu2020optimised, reichel2024LabChip}. Hyperbolic microchannels have emerged as a robust platform for this purpose, providing a controlled hydrodynamic environment to probe continuous cellular responses under progressive mechanical stress \cite{piergiovanni2020LabChip, liu2020optimised, reichel2024LabChip}. Unlike parabolic shear profiles in straight channels, the converging geometry of a hyperbolic channel imposes a nearly constant extensional strain rate along the centerline, driving reproducible cellular deformation trajectories. As cells traverse distinct hydrodynamic regions—specifically the contraction, throat, and expansion (release) phases—their morphology, optical density, and kinematics vary continuously \cite{housiadas2024JFM, chen2022SmartMed}. Tracking these time-resolved profiles generates a complex, high-dimensional feature space. This space encompasses time-dependent, phase-specific, and statistical metrics, offering a comprehensive representation of the cellular mechanophenotype (Figure \ref{fig}). Unlike isolated static snapshots, these dynamic trajectories capture phase-dependent information across discrete deformation and recovery stages.

However, the substantial increase in descriptive dimensionality introduces challenges for reliable image-driven phenotyping. Expanding the feature space does not inherently improve analytical discrimination \cite{negovetic2024efficiently, schneider2025enhancing}. High-dimensional data frequently introduces feature redundancy and increases the risk of algorithmic overfitting, complicating the identification of analytically effective variables. More critically, high classification accuracy under a single experimental condition does not guarantee feature stability. This limitation is particularly evident in dynamic microfluidic phenotyping, where the measured cellular trajectory is jointly shaped by intrinsic cellular properties and the hydrodynamic loading imposed by the operating flow rate. 

Fluctuations in operational parameters, such as volumetric flow rate, directly alter the local hydrodynamic stress distributions experienced by the cells \cite{piergiovanni2020LabChip}. Consequently, dynamic trajectory-derived feature distributions can become mathematically coupled to the underlying hydrodynamic conditions rather than intrinsic cellular biology, introducing classification artifacts \cite{housiadas2024JFM, chen2022SmartMed}. Such coupling shifts feature distributions, distorts the global organization of the feature space, and generates apparent discriminative boundaries that are valid only under specific flow conditions. Without a systematic framework to decouple stable discriminative structures from flow-induced variations, the analytical reliability of high-dimensional dynamic features remains limited. Robust dynamic phenotyping therefore requires feature-level evaluation across flow rates to distinguish reproducible biological structures from flow-conditioned physical artifacts.

To address these limitations, we present an analytical framework designed to extract and isolate hydrodynamically robust features from continuous cellular trajectories in a hyperbolic microchannel. Using a panel of healthy and malignant ovarian cell lines, transient cellular responses are segmented into discrete functional stages. By integrating stage-specific measurements of morphology, kinematics, and intracellular optical density, we construct a high-dimensional feature space that captures dynamic phenotypic behaviors beyond the scope of conventional quasi-static descriptors. To decouple intrinsic biophysical signatures from flow-induced artifacts, inter-class distributional differences are rigorously evaluated across multiple hydrodynamic conditions. Rather than evaluating feature relevance at each flow rate independently, this strategy defines stability at the level of cross-flow discriminative relationships. Features are prioritized when their inter-class effect-size structures remain reproducible across flow rates, mitigating the selection of descriptors whose apparent discriminative value relies on a single flow condition. The analytical stability of this optimized subset is subsequently investigated via principal component analysis and validated across diverse machine learning architectures for both binary (healthy versus malignant) and multi-class (cancer subtyping) classification tasks, establishing a basis for flow-independent, label-free cell profiling.

%----------------------------------------------------------%
\section{Materials and Methods}
\label{sec:Matethod}
%----------------------------------------------------------%

%----------------------------------------------------------%
\subsection{Microfluidic Experiments}
\label{subsec:Exp}
%----------------------------------------------------------%

%--------------------------------------------------%
\begin{figure*}[t]
\centering
\includegraphics[width=\textwidth]{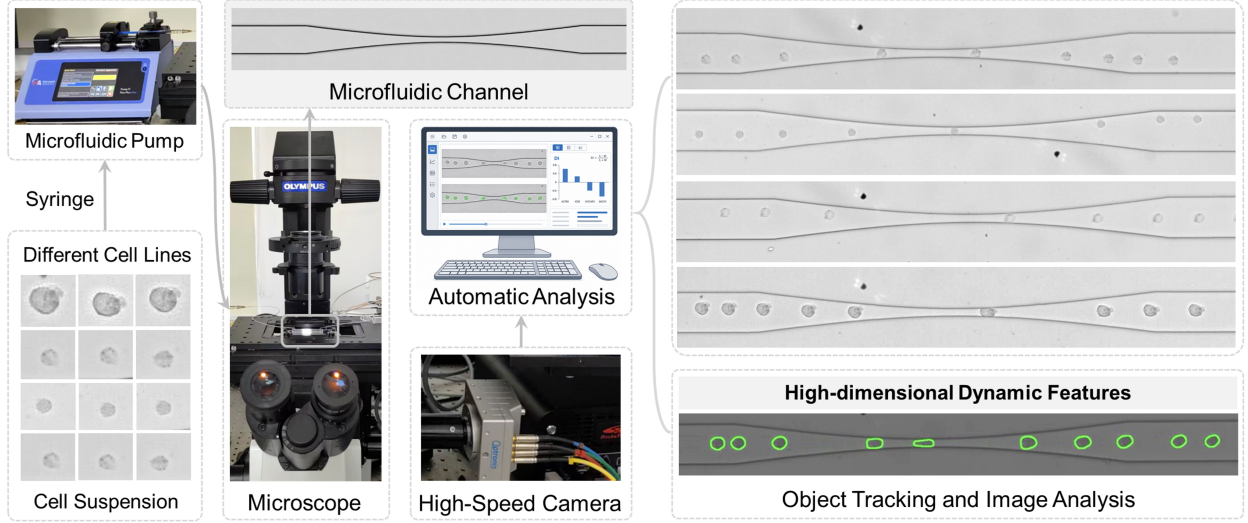}
\caption{Experimental workflow for single-cell transit imaging and dynamic feature extraction in a hyperbolic microfluidic channel. Cell suspensions are introduced via a microfluidic pump, recorded using high-speed bright-field imaging during channel transit, and processed through automated object tracking to extract high-dimensional dynamic features.}
\label{fig:ExpDemo}
\end{figure*}
%--------------------------------------------------%

Four human ovarian-derived cell lines were utilized in the microfluidic experiments. IOSE-80 served as a non-malignant ovarian epithelial control, whereas A2780, SKOV-3, and OVCAR-3 were employed as malignant ovarian cancer lines. This panel spans a broad range of cell sizes and morphological profiles, providing a representative model for evaluating single-cell transport and deformation under confined extensional flow.

%--------------------------------------------------% 
\begin{table}[htbp]
\centering
\caption{Distribution of measured cell samples across evaluated flow rates.}
\label{tab:flow_distribution}
\begin{tabular}{lccccc}
\hline
\textbf{Flow rate} & \textbf{IOSE-80} & \textbf{SKOV-3} & \textbf{OVCAR-3} & \textbf{A2780} & \textbf{Total} \\
\hline
10 $\mu$L/h & 101 & 56  & 101 & 176 & 434 \\
20 $\mu$L/h & 179 & 202 & 182 & 234 & 797 \\
30 $\mu$L/h & 263 & 253 & 160 & 174 & 850 \\
\hline
\textbf{Total} & \textbf{543} & \textbf{511} & \textbf{443} & \textbf{584} & \textbf{2081} \\
\hline
\end{tabular}
\end{table}
%--------------------------------------------------%

Cells were cultured at 37~$^\circ$C in a humidified atmosphere containing 5\% CO$_2$. SKOV-3 cells were maintained in McCoy's 5A medium; IOSE-80, A2780, and OVCAR-3 cells were maintained in RPMI 1640 medium. Media were supplemented with 10\% fetal bovine serum and 1\% penicillin--streptomycin. Prior to microfluidic measurements, cells were harvested during the logarithmic growth phase and resuspended at a final concentration of $1 \times 10^6$~cells/mL to ensure isolated single-cell transit events.

To minimize gravitational sedimentation during fluidic transport, cells were suspended in a density-adjusted medium prepared by adding 20\% (v/v) OptiPrep to the basal culture medium, achieving a final suspension density of approximately 1.05~g/cm$^3$. This adjustment maintained cells near the imaging focal plane, improving recording consistency.

The microfluidic device consisted of a symmetric hyperbolic contraction--expansion channel flanked by straight regions. The central throat possessed a half-width of 6~$\mu$m, and the channel depth was uniformly maintained at 15~$\mu$m. This geometry exposes cells to progressive spatial confinement during entry, maximum extensional deformation at the throat, and viscoelastic recovery in the expansion region. Devices were fabricated via standard soft lithography. Briefly, polydimethylsiloxane (PDMS) was cast over a patterned SU-8 master mold at a 10:1 base-to-curing-agent ratio. Following curing at 80~$^\circ$C, the PDMS replicas were bonded to glass substrates via oxygen-plasma treatment. Prior to experimentation, channels were passivated with 1\% (w/v) Pluronic F127 for 45~min to mitigate nonspecific cellular adhesion. Microfluidic measurements were conducted using a syringe pump at volumetric flow rates of 10, 20, and 30~$\mu$L/h. Bright-field image sequences of single-cell transit were acquired using an inverted microscope coupled to a high-speed camera once flow stabilized.

%----------------------------------------------------------%
\subsection{Automated Image Processing and Dynamic Feature Database}
%----------------------------------------------------------%

A custom OpenCV-based pipeline was implemented to extract quantitative frame-wise descriptors from the raw image sequences. To account for batch-to-batch illumination variations, a spatial and intensity normalization protocol was applied. A cell-free reference frame from each sequence was converted to 8-bit grayscale to define a constant region of interest (ROI). The background baseline, $I_{bg}$, was calculated as:
\begin{equation}
    I_{bg} = \frac{1}{N_{ROI}} \sum_{i \in ROI} I(i)
\end{equation}
Subsequent pixel intensities were normalized as $I'_i = I(i) / I_{bg}$, establishing a uniform baseline for intensity-derived metrics.

Continuous cell contours ($C_t$) and segmented areas ($\Omega_t$) were extracted frame-by-frame via background subtraction, binarization, and morphological filtering (Supporting Information). Spatially, contour coordinates were translated relative to the geometric center of the channel ($x_0, y_0$), yielding $C'_{x,t} = C_{x,t} - x_0$ and $C'_{y,t} = C_{y,t} - y_0$. This coordinate normalization established a unified spatial reference frame, ensuring the comparability of cellular trajectories across experimental conditions. Events involving multiple cells, cellular debris, or boundary truncation were excluded. Based on these validated masks, primary quasi-static measurements were extracted per frame: length ($L_t$), width ($W_t$), perimeter ($P_t$), area ($A_t$), mean optical intensity ($MeanGray_t$)~\cite{carpenter2006cellprofiler}, and centroid velocity ($V_t$). 

To characterize deformation-associated dynamic responses, primary measurements were algebraically combined to derive functional morphodynamic variables: Aspect Ratio ($AR_t = L_t / W_t$), Roundness ($R_t = 4\pi A_t / P_t^2$), and the Taylor deformation parameter ($D_t = (L_t - W_t)/(L_t + W_t)$). Cellular trajectories were spatially segmented into four functional stages based on the centroid position along the channel axis: entrance (E), contraction (C), release (R), and exit (X). Dynamic measurements extracted from these temporal trajectories were organized into a structured dynamic feature space comprising four analytical representations:

\begin{itemize}
\item \textbf{Stage-specific statistical descriptors:} Mean, variance, and coefficient of variation of the morphodynamic variables within each functional stage.
\item \textbf{Inter-stage transition descriptors:} Differential metrics quantifying the magnitude of structural transitions between adjacent deformation stages.
\item \textbf{Trajectory-dynamic descriptors:} Kinetic parameters characterizing release-region recovery dynamics, with relaxation time constants estimated via exponential fitting of velocity and shape recovery profiles~\cite{fregin2019high,reichel2024LabChip}.
\item \textbf{Global response descriptors:} Aggregate kinematic descriptors summarizing the overall transit behavior across the complete trajectory (e.g., mean velocity, peak velocity).
\end{itemize}
The complete definitions and nomenclature of all variables included in the full dynamic feature space (93 features) are detailed in Table~S1 (Supporting Information).

%----------------------------------------------------------%
\subsection{Task-Driven Feature Analysis and Statistical Framework}
%----------------------------------------------------------%

Cellular mechanophenotyping typically addresses two progressive analytical objectives: differentiating malignant cells from healthy epithelial counterparts (binary classification, hereafter designated as Phenotype 1) and distinguishing among specific malignant clonal lineages (multi-class subtyping, Phenotype 2). To evaluate the representational capacity of the extracted features across these diagnostic tasks, statistical assessments were structured into a progressive framework: (1) Descriptor-Specific Subspace Construction; (2) Principal Component Analysis and Variance Partitioning; (3) Flow-Induced Stability Analysis; and (4) Classification-Based Feature Validation.

%------------------------------------------------------%
\subsubsection{Descriptor-Specific Subspace Construction}
\label{sec:FeatureSubspace}
%------------------------------------------------------%

Conventional morphological analyses frequently integrate highly correlated geometric descriptors (e.g., roundness, aspect ratio, and the Taylor parameter) within a single predictive model. This redundancy introduces mathematical cross-talk and elevates the risk of algorithmic overfitting. To isolate the intrinsic discriminative contribution of distinct morphodynamic aspects, the global dynamic feature database was partitioned into three structurally equivalent subspaces centered on Roundness, Aspect Ratio, and Taylor parameters. These subspaces were constructed using identical stage-wise statistical encoding strategies across the functional microchannel zones. This symmetrical construction ensures that observed performance variations reflect the biophysical relevance of the core descriptor rather than discrepancies in dimensionality. The detailed composition of these subspaces is summarized in Table~S2.

Descriptor-specific subspaces were evaluated independently for the Phenotype 1 and Phenotype 2 tasks using five-times-repeated five-fold stratified cross-validation. Paired $t$-tests were performed across the cross-validation results to evaluate the statistical significance of performance discrepancies. The $t$-statistic is defined as:
\begin{equation}
    t = \frac{\bar{d} - \mu_0}{s_d / \sqrt{n}}
\end{equation}
where $\bar{d}$ is the sample mean of the paired performance differences, $\mu_0$ is the hypothesized mean difference ($0$), $s_d$ is the sample standard deviation, and $n$ is the total number of paired evaluation results. 

To mitigate analytical biases introduced by class imbalance, the Macro-F1 score was selected as the primary performance metric:
\begin{equation}
    \text{Macro-F1} = \frac{1}{C} \sum_{c=1}^{C} \frac{2 P_c R_c}{P_c+R_c}
\end{equation}
where $C$ denotes the total number of classes, and $P_c$ and $R_c$ represent precision and recall for class $c$, respectively.

%------------------------------------------------------%
\subsubsection{Principal Component Analysis and Variance Partitioning}
%------------------------------------------------------%

Principal Component Analysis (PCA) was utilized to characterize the global organization of the dynamic feature space. All variables were standardized via z-score normalization across the pooled dataset containing all evaluated flow rates. PCA was conducted without stratification by flow rate to evaluate the relative contributions of hydrodynamics and phenotypic class to the overall feature-space topology.

To quantify these relative contributions, an ANOVA-based variance-attribution analysis was applied to the primary principal component scores (PC1 and PC2)~\cite{bertinetto2020anova}. Linear models were constructed using flow rate and phenotypic class labels as categorical variables:
\begin{equation}
    PC_i \sim C(\mathrm{flow\ rate}) + C(\mathrm{phenotypic\ class})
\end{equation}
where $PC_i$ denotes the score of the $i$-th principal component. The sums of squares associated with each experimental factor were calculated using type-II ANOVA and expressed as relative variance contributions.

%------------------------------------------------------%
\subsubsection{Flow-Induced Stability Analysis}
%------------------------------------------------------%

Because hydrodynamic perturbations alter cellular deformation responses \cite{piergiovanni2020LabChip,reichel2024LabChip}, preserving discriminative relationships across flow rates is essential for defining robust mechanophenotypic markers. Rather than comparing heterogeneous variables with differing numerical scales directly, all candidate descriptors were transformed into a common effect-size representation using Cliff's delta ($\delta$)~\cite{masuda2021integrative}. A candidate subset ($S_{\text{candidate}}$) was constructed using a dual-level screening strategy based on structural consistency and statistical persistence across flow rates.

For the binary classification task (Phenotype 1), observations from all malignant lineages were aggregated into a unified cancer cohort. Structural consistency requires that the directional polarity of the phenotypic shift remains invariant across all hydrodynamic conditions:
\begin{equation}
    \operatorname{sgn}\left( \delta_{\text{binary}}^{(v)} \right) = \operatorname{sgn}\left( \delta_{\text{binary}}^{(v')} \right), \quad \forall v, v' \in V
\end{equation}
where $\operatorname{sgn}(\cdot)$ denotes the signum function, $\delta_{\text{binary}}^{(v)}$ represents the inter-class effect size at flow rate $v$, and $V$ denotes the complete set of operational flow conditions. The strict preservation of this effect-size polarity ensures the target feature captures an intrinsic biophysical divergence rather than a transient, flow-induced artifact.

For the multi-class subtyping task (Phenotype 2), pairwise Cliff's delta values were calculated independently for each subtype combination (e.g., A2780 vs. SKOV-3, denoted as A-S). Structural consistency in the multi-class context is dictated by the preservation of the relative discriminative hierarchy rather than simple directional polarity. This requires the ordinal ranking of the absolute effect sizes to remain strictly invariant across all evaluated flow rates (e.g., $\text{A-S} > \text{O-S} > \text{A-O}$). 

In parallel, statistical persistence was strictly defined as the bootstrap 95\% confidence intervals of $\delta$ (estimated via 1,000 resampling iterations) completely excluding zero across all evaluated flow rates. Finally, a dual-threshold strategy was applied to ensure candidate features maintained globally sufficient discriminative strength. This was mathematically enforced by requiring both the cross-condition mean absolute effect size and the condition-specific minimum absolute effect size to exceed their respective global baselines:
\begin{equation}
    \frac{1}{N_V} \sum_{v \in V} |\delta^{(v)}(\text{feature x})| > \theta_{\text{mean}}
\end{equation}
\begin{equation}
    \min_{v \in V} |\delta^{(v)}(\text{feature x})| > \theta_{\text{min}}
\end{equation}
where $N_V$ represents the total number of evaluated flow rates, and $\theta_{\text{mean}}$ and $\theta_{\text{min}}$ denote the global averages of the mean and minimum absolute effect sizes computed across the entire feature space. Features satisfying both the stability criteria and this dual-intensity constraint formed the optimized final subset ($S_{\text{final}}$).

%------------------------------------------------------%
\subsubsection{Classification-Based Feature Validation}
%------------------------------------------------------%

To verify that the stable discriminative structures preserved in $S_{\text{final}}$ translate into practical classification performance, probability-distribution-based separability analysis and supervised classification validation were performed.

A classifier-independent composite separability score ($S_{\mathrm{comp}}$) was formulated. Class probabilities were generated using a Support Vector Machine (SVM) during repeated stratified cross-validation. For binary classification, the predicted probability was transformed into a logit score $z=\log(p/(1-p))$. The resulting logit-score distributions were compared using the Bhattacharyya distance ($D_B$), symmetric Kullback--Leibler divergence ($D_{SKL}$), and Fisher ratio ($FR$). Continuous scores were discretized into shared histogram probability densities ($p_i, q_i$):
\begin{equation}
\begin{aligned}
D_B &= -\ln \left( \sum_i \sqrt{p_iq_i} \right), \\
D_{SKL} &= \frac{1}{2} \left[ \sum_i p_i \ln \frac{p_i}{q_i} + \sum_i q_i \ln \frac{q_i}{p_i} \right], \\
FR &= \frac{(\mu_p-\mu_q)^2}{\sigma_p^2+\sigma_q^2}
\end{aligned}
\end{equation}
These complementary measures were aggregated using an equal-weight strategy~\cite{joint2008handbook}:
\begin{equation}
S_{\mathrm{comp}} = D_B + D_{SKL} + \ln(1+FR)
\end{equation}
Higher $S_{\mathrm{comp}}$ values indicate stronger discriminative organization in the probability-score space. For multi-class analysis, a one-versus-rest strategy was adopted, and class-specific scores were averaged. 

To evaluate practical utility, validation was performed across multiple supervised models: SVM, Logistic Regression (LR), Random Forest (RF), and K-Nearest Neighbors (KNN)~\cite{nitta2018intelligent,ciaparrone2024label}. Model robustness was assessed under progressively varying training-to-testing split ratios (50:50 to 90:10) using repeated stratified shuffle-split validation at an intermediate flow rate (20~$\mu$L/h).

%---------------------------------------------------%
\section{Results and Discussion}
%---------------------------------------------------%

%---------------------------------------------------%
\subsection{Comparative Subspace Evaluation and Task-Dependent Descriptors}
%---------------------------------------------------%

To determine whether distinct diagnostic granularities preferentially exploit specific morphodynamic signatures, the three descriptor-specific subspaces (Roundness, Aspect Ratio, and Taylor parameter) were evaluated independently. 

\begin{figure}[!thpb]
    \centering
    \begin{minipage}[t]{0.475\textwidth}
    \vspace{0pt}
    \centering
    \begin{overpic}[width=\linewidth]{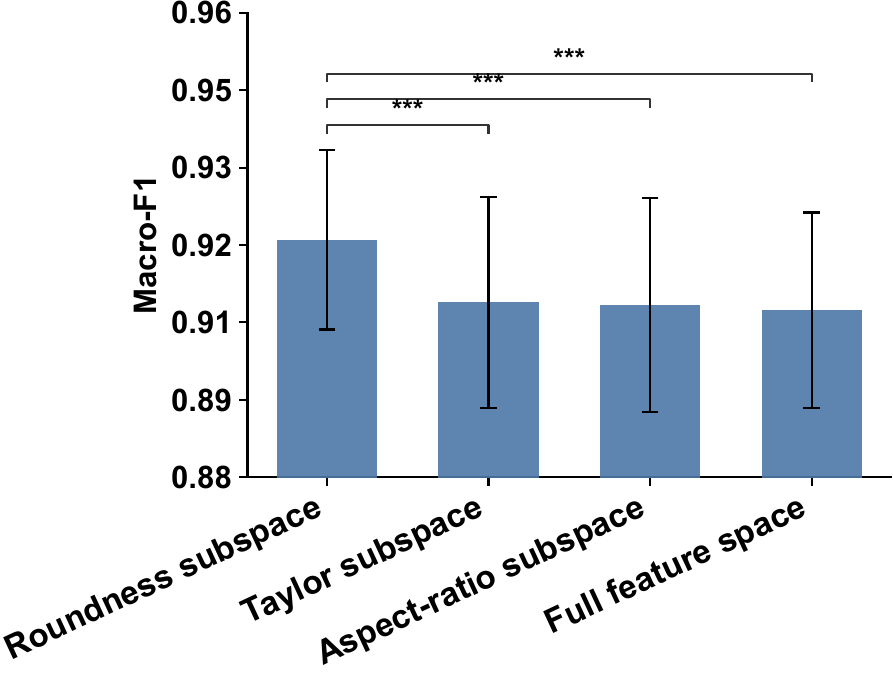}
    \put(0,75){\large (a)}
    \end{overpic}
    \end{minipage}
    %-------------------------------------------%
    \hspace{0.015\textwidth}
    \begin{minipage}[t]{0.475\textwidth}
    \vspace{0pt}
    \centering
    \begin{overpic}[width=\linewidth]{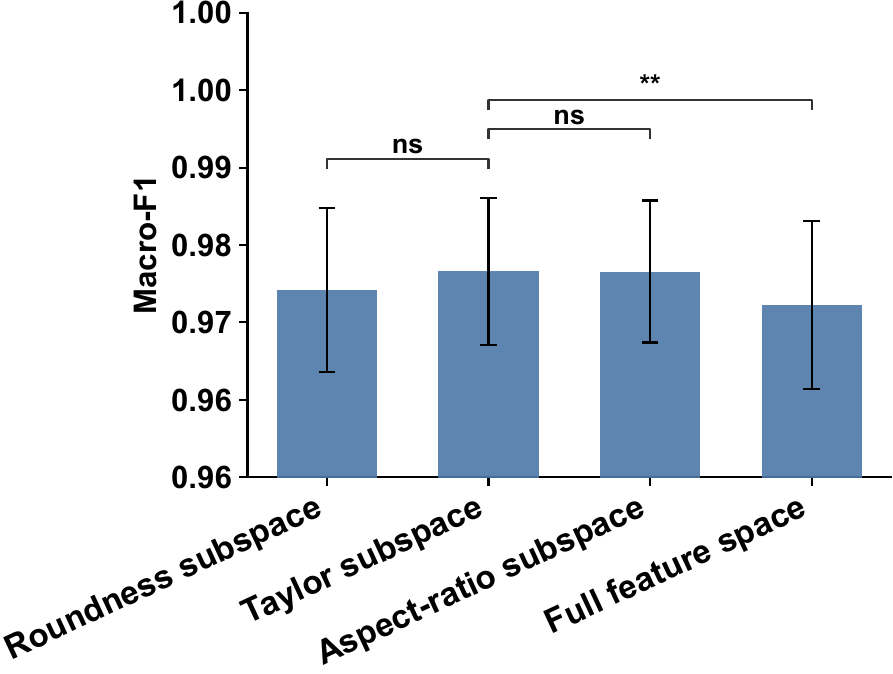}
    \put(0,75){\large (b)}
    \end{overpic}
    \end{minipage}
    \caption{Performance comparison of descriptor-specific dynamic subspaces. (a) Phenotype 1, cancer-versus-healthy binary classification. (b) Phenotype 2, cancer-subtype multi-class classification. Significance levels: \textsuperscript{ns}, $p \geq 0.05$; \textsuperscript{**}, $p < 0.01$; \textsuperscript{***}, $p < 0.001$.}
    \label{fig:subspace_performance}
\end{figure}

For the binary classification task (healthy versus malignant epithelial cells, Phenotype 1), paired $t$-tests across a five-times-repeated, 5-fold stratified cross-validation indicated that the Roundness-related subspace significantly outperformed both the global 93-dimensional feature space and the alternative subspaces (all $p < 0.001$). Evaluation based on ROC-AUC mirrored this trend, indicating that the Roundness subspace preserves the most robust overall separation boundary between healthy and malignant states (Table~\ref{tab:subspace_performance} and Figure~\ref{fig:subspace_performance}a). These observations imply that the macroscopic transition from a healthy to a malignant state is predominantly captured by bulk morphological irregularities.

Conversely, the descriptor-specific subspaces exhibited closer performance margins in the multi-class cancer-subtype classification (Phenotype 2, Figure~\ref{fig:subspace_performance}b). The Taylor-related subspace yielded the highest discriminative performance, significantly outperforming the global dynamic space under paired comparison ($p < 0.01$). Evaluation based on ROC-AUC similarly demonstrated that the Taylor subspace attained the highest metric. Because the Taylor parameter directly quantifies active extensional cell strain, it provides sensitivity to subtle viscoelastic variations inherent to distinct cell lineages. Such active deformation metrics isolate resistance to extensional stress, which is often dictated by the structural integrity of the actin cortex and nuclear envelope—features known to vary among distinct clonal lineages.

Collectively, these findings demonstrate that different diagnostic tasks necessitate distinct physical representations. Binary discrimination is primarily governed by bulk morphodynamic boundaries captured by Roundness-related descriptors, whereas cancer subtyping requires the high-resolution characterization of active deformation kinematics afforded by Taylor-related parameters.

%-----------------------------------------------------%
\begin{table}[!thpb]
\centering
\caption{Classification performance of the full feature space and descriptor-specific dynamic subspaces for Phenotypes 1 and 2. Detailed feature descriptions are provided in Supplementary Tables~S1 and S2.}
\begin{tabular}{llcll}
\hline
\textbf{Task} & \textbf{Feature Subspace} & \textbf{Feature Count} & \textbf{Macro-F1} & \textbf{ROC-AUC} \\
\hline
Phenotype 1 & Full feature space & 93 & 0.9091 ± 0.0164 & 0.9739 ± 0.0054 \\
Phenotype 1 & Roundness subspace & 77 & 0.9209 ± 0.0151 & 0.9780 ± 0.0062 \\
Phenotype 1 & Taylor subspace & 77 & 0.9104 ± 0.0178 & 0.9766 ± 0.0052 \\
Phenotype 1 & Aspect-ratio subspace & 77 & 0.9099 ± 0.0180 & 0.9762 ± 0.0052 \\
Phenotype 2 & Full feature space & 93 & 0.9742 ± 0.0085 & 0.9976 ± 0.0013 \\
Phenotype 2 & Roundness subspace & 77 & 0.9757 ± 0.0083 & 0.9978 ± 0.0011 \\
Phenotype 2 & Taylor subspace & 77 & 0.9775 ± 0.0075 & 0.9978 ± 0.0011 \\
Phenotype 2 & Aspect-ratio subspace & 77 & 0.9775 ± 0.0072 & 0.9978 ± 0.0011 \\
\hline
\end{tabular}
\label{tab:subspace_performance}
\end{table}
%-----------------------------------------------------%

%-----------------------------------------------------%
%-----------------------------------------------------%

%---------------------------------------------------%
\subsection{Stability-Guided Feature Selection and Composition Mapping}
%---------------------------------------------------%

\begin{figure*}[!thpb]
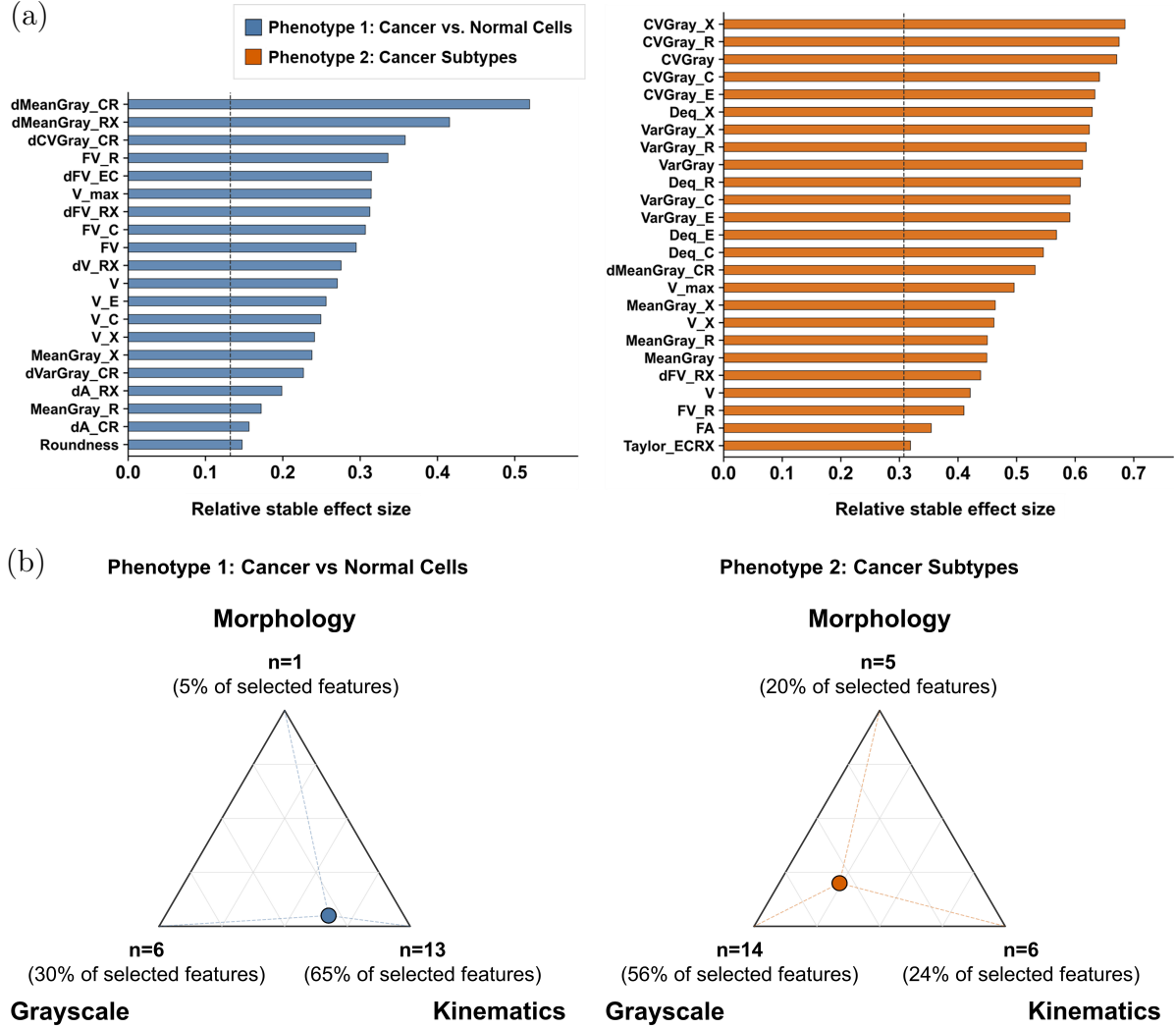

    \centering
    %-------------------------------------------%
    \begin{minipage}[h]{0.95\textwidth}
    \vspace{10pt}
    \centering
    \begin{overpic}[width=\linewidth]{Figure4e.pdf}
    \put(0,42.5){\large (a)}
    \end{overpic}
    \end{minipage}
    %-------------------------------------------%
    \vspace{0.15em}
    %-------------------------------------------%
    \begin{minipage}[h]{0.95\textwidth}
    \vspace{10pt}
    \centering
    \begin{overpic}[width=\linewidth]{Figure4g.pdf}
    \put(0,40){\large (b)}
    \end{overpic}
    \end{minipage}
    %-------------------------------------------%
    \vspace{0.2em}
    %-------------------------------------------%
\caption{Feature-level overview of stability-guided refinement. (a) Effect-size ranking of the retained features after cross-flow stability screening. Bars show the relative stable effect size, calculated from the cross-flow mean absolute Cliff's delta after subtraction of the minimum-effect baseline. Dashed lines indicate task-specific mean-effect thresholds. (b) Ternary maps of the retained feature subsets according to morphology, grayscale, and kinematics. The centroid represents the count-based proportions within each refined subset.}
\label{fig:stable_contribute}
\end{figure*}

Following subspace validation, non-parametric stability screening via Cliff's delta was applied to isolate individual features that preserve inter-class discriminative relationships across operational flow variations. This framework compressed the feature space, retaining 20 variables for the binary task (Phenotype 1) and 25 variables for the multi-class subtyping task (Phenotype 2), as shown in Figure~\ref{fig:stable_contribute}a. The retained features map across a broad spectrum of stable effect sizes rather than clustering artificially near threshold boundaries, indicating that the protocol successfully preserves multiple distinct features exhibiting robust discriminative capacity. Such a stability-oriented rationale is consistent with contemporary biomarker-discovery frameworks that prioritize reproducibility in high-dimensional screening \cite{luo2025stability,liu2025stable}.

Mapping the physical composition of these refined subsets reveals a fundamental shift in the mechanophenotypic strategy contingent upon diagnostic granularity (Figure~\ref{fig:stable_contribute}b). The binary subset is predominantly composed of kinematic descriptors (65.0\%), alongside a secondary grayscale-associated component (30.0\%). In contrast, the multi-class subset relies heavily on grayscale-associated features (56.0\%), supplemented by comparatively smaller contributions from kinematics (24.0\%) and morphodynamics (20.0\%). Grayscale profiles in bright-field imaging correlate with intracellular refractive index gradients, reflecting localized variations in organelle packing and chromatin condensation. This compositional shift demonstrates that as diagnostic resolution increases, accurate subtyping requires profiling highly localized internal optical heterogeneity rather than macroscopic kinematic responses alone.

%========================================================%
%========================================================%

\begin{figure*}[!t]
    \centering    
    \begin{minipage}[t]{0.475\textwidth}
    \centering
    \vspace{0pt}
    \begin{overpic}[width=\linewidth]{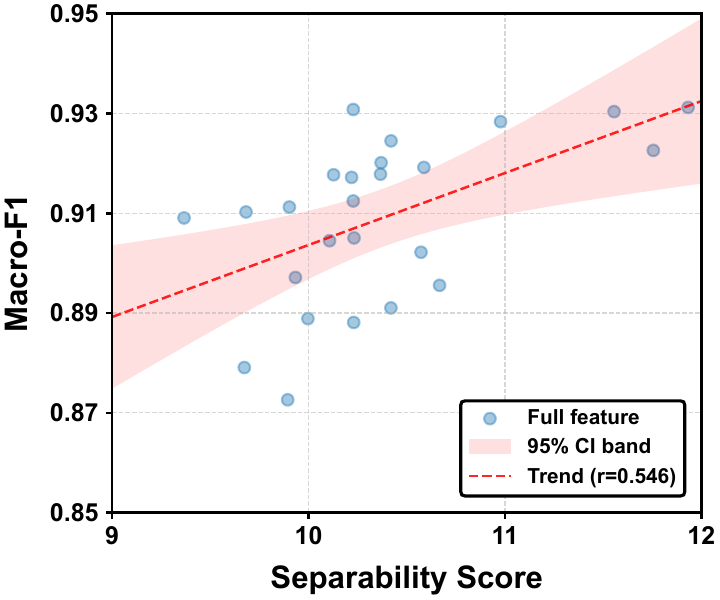}
    \put(-0.5,80){\large{(a)}}
    \end{overpic}
    \end{minipage}
    %------------------------------------------------%
    \begin{minipage}[t]{0.475\textwidth}
    \centering
    \vspace{0pt}
    \begin{overpic}[width=\linewidth]{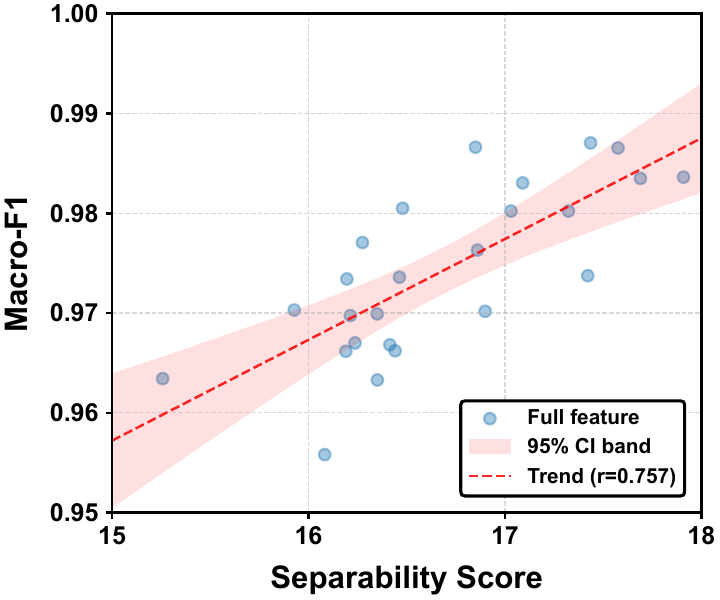}
    \put(-0.5,80){\large{(b)}}
    \end{overpic}
    \end{minipage}
    %------------------------------------------------%
    \begin{minipage}[t]{0.475\textwidth}
    \centering
    \vspace{10pt}
    \begin{overpic}[width=\linewidth]{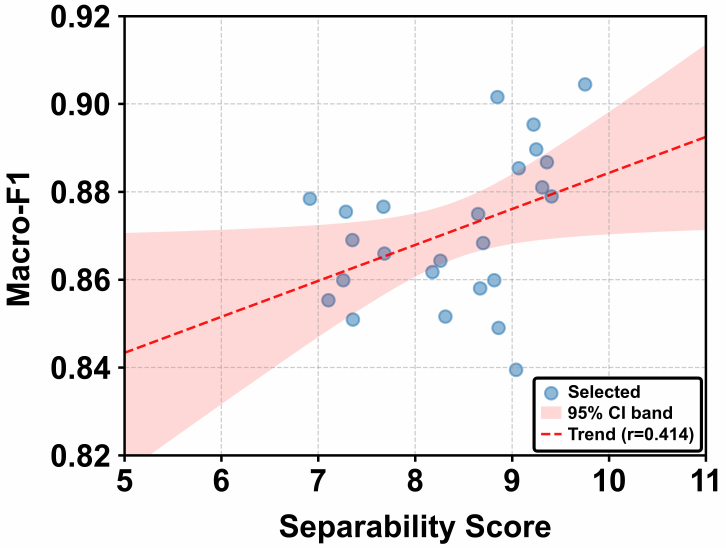}
    \put(-0.5,75){\large{(c)}}
    \end{overpic}
    \end{minipage}
    %------------------------------------------------%
    \begin{minipage}[t]{0.475\textwidth}
    \centering
    \vspace{10pt}
    \begin{overpic}[width=\linewidth]{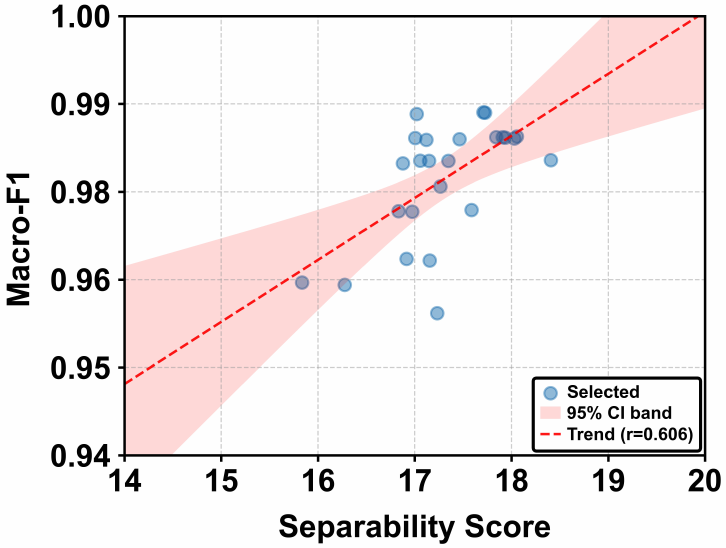}
    \put(-0.5,75){\large{(d)}}
    \end{overpic}
    \end{minipage}
    %------------------------------------------------%
\caption{Separability--performance relationships before and after stability-guided feature refinement. (a, b) Fold-level correlations between composite separability score and Macro-F1 in the task-adapted Roundness and Taylor subspaces. (c, d) Corresponding correlations in the retained feature subsets after stability-guided refinement. Dashed lines indicate linear trends, and shaded regions denote 95\% confidence bands.}
\label{fig:F1vsSeparab}
\end{figure*}

To verify that the overarching discriminative topology was preserved following feature reduction, classifier-independent separability analysis was performed (Figure~\ref{fig:F1vsSeparab}). The correlation between the composite separability score and the downstream Macro-F1 score establishes an analytical baseline for the unrefined features in the binary (Figure~\ref{fig:F1vsSeparab}a) and multi-class (Figure~\ref{fig:F1vsSeparab}b) tasks. Positive Pearson correlations were consistently maintained across this analytical progression: from the initial macroscopic biological subspaces (Roundness-related binary, $r=0.606$; Taylor-related multi-class, $r=0.757$) to the final stability-screened subsets (binary, $r=0.414$; multi-class, $r=0.606$). In all instances, cross-validation folds demonstrating stronger mathematical feature-space separation achieved correspondingly higher Macro-F1 values. These reproducible trends confirm that the fundamental discriminative architecture identified within the original high-dimensional manifolds remains mathematically intact following dimensional compression.

%========================================================%
%========================================================%

%---------------------------------------------------%
\subsection{Decoupling Hydrodynamic Artifacts from Biological Signatures}
%---------------------------------------------------%

To systematically evaluate the sensitivity of the extracted feature spaces to flow-induced artifacts, PCA combined with ANOVA-based variance partitioning, alongside cross-condition structural heatmaps, were applied to quantify the structural topology.

\begin{figure*}[!t]
\centering
    %--------------------------------------%
    \begin{overpic}[width=0.455\textwidth]{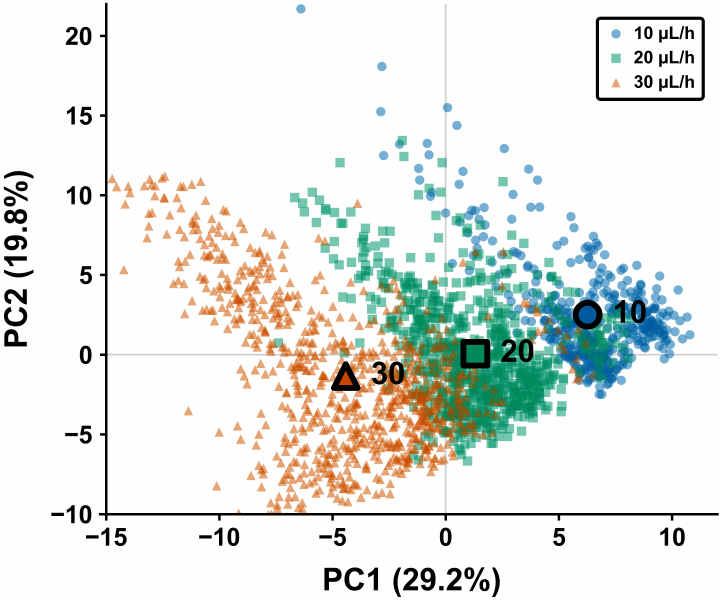}
    \put(-2,80){\large{(a)}}
    \end{overpic}
    \hfill
    \begin{overpic}[width=0.455\textwidth]{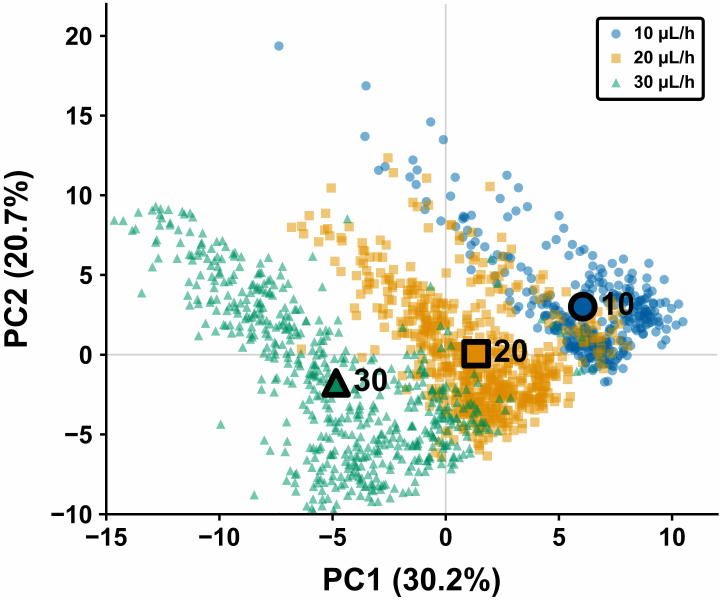}
    \put(-2,80){\large{(b)}}
    \end{overpic}
    %--------------------------------------%
    \begin{overpic}[width=0.455\textwidth]{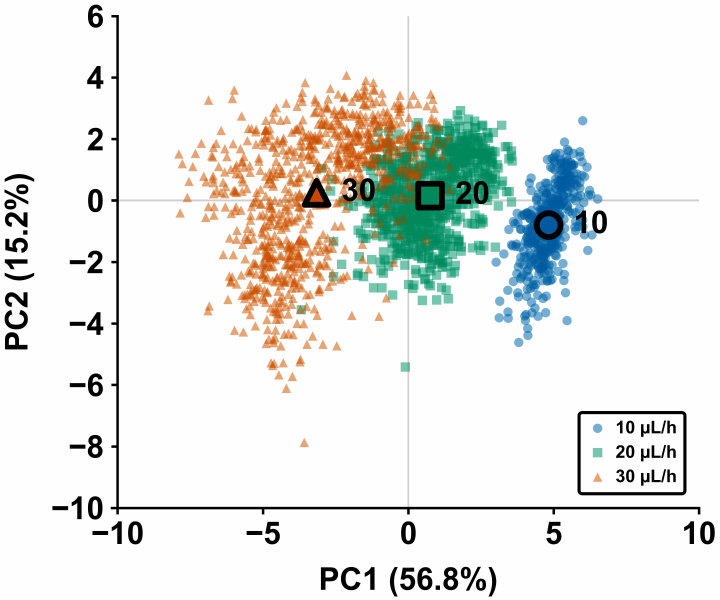}
    \put(0,80){\large{(c)}}
    \end{overpic}
    \hfill
    \begin{overpic}[width=0.455\textwidth]{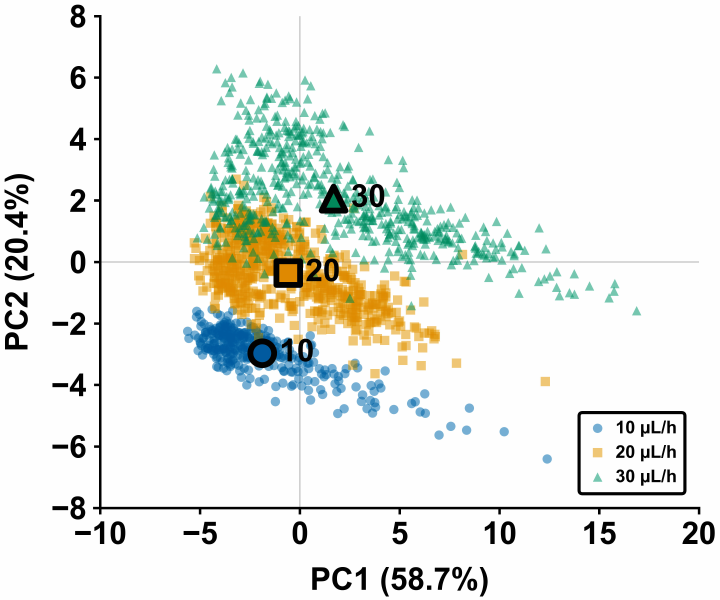}
    \put(0,80){\large{(d)}}
    \end{overpic}
    %--------------------------------------%
\caption{PCA-based evaluation of flow-conditioned feature-space organization before and after stability-guided refinement. (a, b) PCA projections of the original dynamic feature spaces for Phenotype 1 and Phenotype 2, respectively, colored by flow rate. (c, d) PCA projections of the corresponding refined feature spaces after stability-guided refinement, colored by flow rate.}
\label{fig:PCA_FlowRate}
\end{figure*}

In the original high-dimensional dynamic space, PCA projections exposed severe hydrodynamic drift; sample centroids migrated systematically across the principal component planes as a direct function of the imposed fluidic velocity (Figures~\ref{fig:PCA_FlowRate}a and \ref{fig:PCA_FlowRate}b). Variance-attribution analysis quantitatively confirmed this artifact, revealing that the primary principal components were mathematically dominated by operational flow conditions rather than phenotypic class. As shown in Table~\ref{tab:pca_contributions}, the flow rate accounted for $94.93\%$ and $99.90\%$ of the explained variance within PC1 and PC2, respectively, in the binary task. Similarly, in the multi-class subtyping task, flow rate remained the dominant contributor to PC1 ($69.87\%$). These observations confirm that unrefined dynamic feature spaces are heavily confounded by flow-induced variations, potentially obscuring intrinsic biophysical differences.

Conversely, projecting the samples onto the stability-guided manifolds effectively neutralized this flow-conditioned spatial migration. The refined PCA projections preserved the dominant class-associated geometry while substantially compressing flow-induced intra-class dispersion (Figures~\ref{fig:PCA_FlowRate}c and \ref{fig:PCA_FlowRate}d). Variance partitioning mathematically confirmed this structural decoupling. In the binary task, PC2 shifted from being almost entirely flow-dominated in the original space to predominantly class-associated following refinement ($64.08\%$ class vs. $35.92\%$ flow in Table~\ref{tab:pca_contributions_refined}). In the multi-class task, PC1 underwent an almost complete mathematical inversion, becoming overwhelmingly class-associated ($90.67\%$) while restricting flow-associated variance to $9.33\%$.

By rigorously penalizing highly flow-variant descriptors via cross-condition evaluations, the selection procedure systematically purges hydrodynamic artifacts and eliminates mathematical cross-talk among redundant variables. Consequently, the refined subsets secure an intrinsically stable discriminative architecture across varying microfluidic operational parameters.

%========================================================%
%========================================================%

%========================================================%
\begin{table}[!thpb]
    \centering
    \caption{Relative variance contributions of flow rate and phenotypic class to the primary principal components in the original dynamic feature spaces.}
    \label{tab:pca_contributions}
    \begin{tabular}{lccc}
    \hline
    \textbf{Task} & \textbf{Principal Component} & \textbf{Flow Rate Contribution (\%)} & \textbf{Class Contribution (\%)} \\
    \hline
    Phenotype 1 & PC1 & 94.9 & 5.1 \\
    Phenotype 1 & PC2 & 99.9 & 0.1 \\
    Phenotype 2 & PC1 & 69.9 & 30.1 \\
    Phenotype 2 & PC2 & 38.7 & 61.3 \\
    \hline
\end{tabular}
\end{table}
%========================================================%

%========================================================%
\begin{table}[htbp]
\centering
\caption{Relative variance contributions of flow rate and phenotypic class to the primary principal components after stability-guided feature refinement.}
\label{tab:pca_contributions_refined}
\begin{tabular}{lccc}
\hline
\textbf{Task} & \textbf{Principal Component} & \textbf{Flow Rate Contribution (\%)} & \textbf{Class Contribution (\%)} \\
\hline
Phenotype 1 & PC1 & 96.1 & 3.9 \\
Phenotype 1 & PC2 & 35.9 & 64.1 \\
Phenotype 2 & PC1 & 9.3 & 90.7 \\
Phenotype 2 & PC2 & 84.3 & 15.7 \\
\hline
\end{tabular}
\end{table}
%========================================================%

\begin{figure*}[!t]
\centering
    %---------------------------------------------------%
    \begin{minipage}[c]{0.625\textwidth}
    \centering
    \begin{overpic}[width=\linewidth]{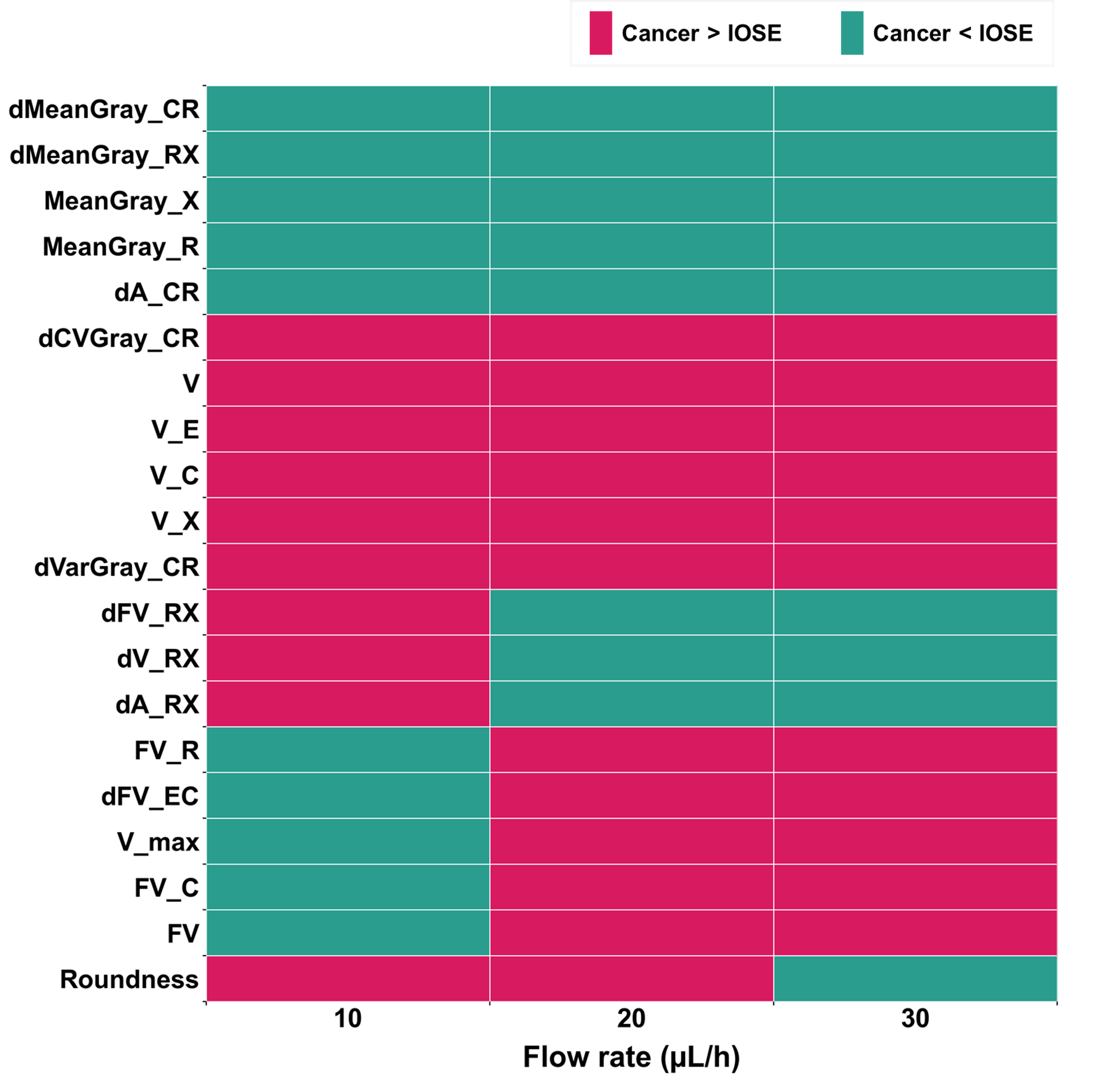}
    \put(0,98){\large (a)}
    \end{overpic}
    \end{minipage}
    \hfill    
    %--------------------------------%
    \begin{minipage}[c]{0.355\textwidth}
    \centering
    %--------------------------------%
    \begin{overpic}[width=\linewidth]{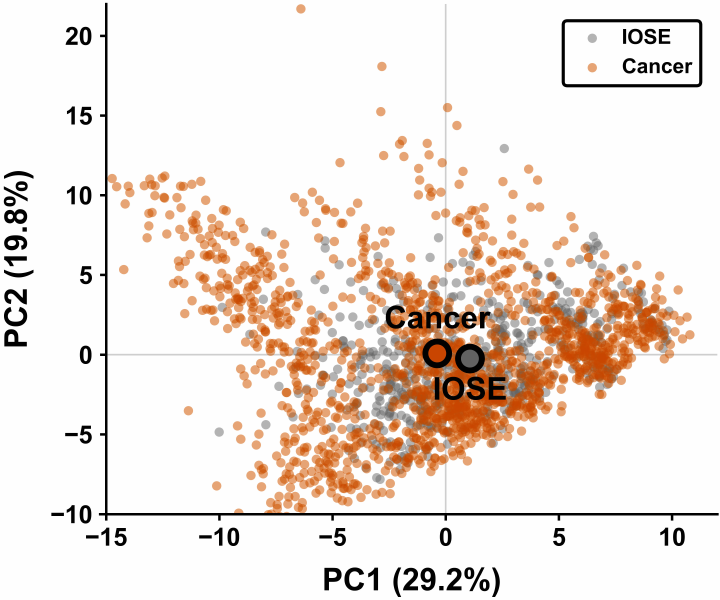}
    \put(-2,80){\large(b)}
    \end{overpic}
    \vspace{0.8em}
    %--------------------------------%
    \begin{overpic}[width=\linewidth]{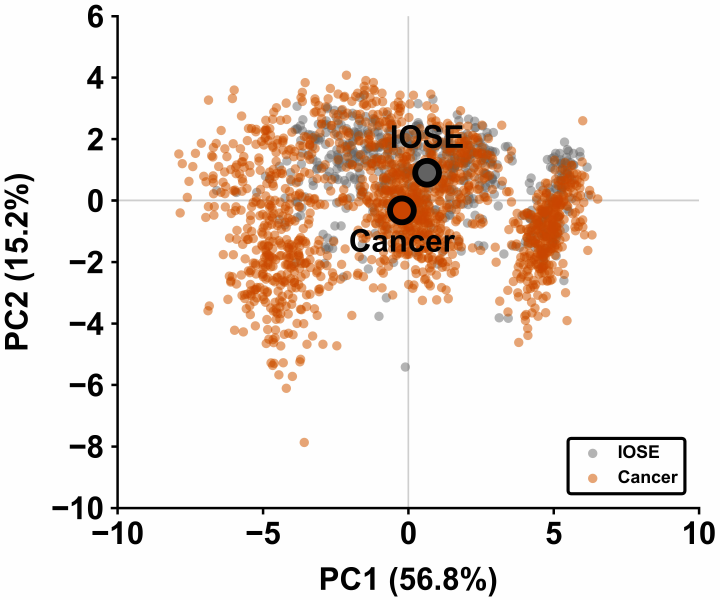}
    \put(-2,80){\large (c)}
    \end{overpic}
    %--------------------------------%    
    \end{minipage}
    %--------------------------------%
\caption{Cross-flow preservation of discriminative structures in the refined Phenotype 1 feature space. (a) Cross-flow structural heatmap of the retained subset, showing the direction of pairwise Cliff's delta across flow rates (10, 20, and 30~$\mu$L/h). Colors indicate whether each retained feature consistently exhibited Cancer $>$ IOSE-80 or Cancer $<$ IOSE-80. (b,c) PCA projections colored by cell type for the original full dynamic feature space and the retained 20-feature subset, respectively. The refined subset preserved the dominant cancer-versus-healthy organization and produced clearer centroid separation.}
\label{fig:HeatMap_Binary}
\end{figure*}

\begin{figure*}[!t]
\centering
    %---------------------------------------------------%
    \begin{minipage}[c]{0.565\textwidth}
    \centering
    \begin{overpic}[width=\linewidth]{Figure6b.pdf}
    \put(0,98){\large (a)}
    \end{overpic}
    \end{minipage}
    \hfill    
    %--------------------------------%
    \begin{minipage}[c]{0.400\textwidth}
    \centering
    %--------------------------------%
    \begin{overpic}[width=\linewidth]{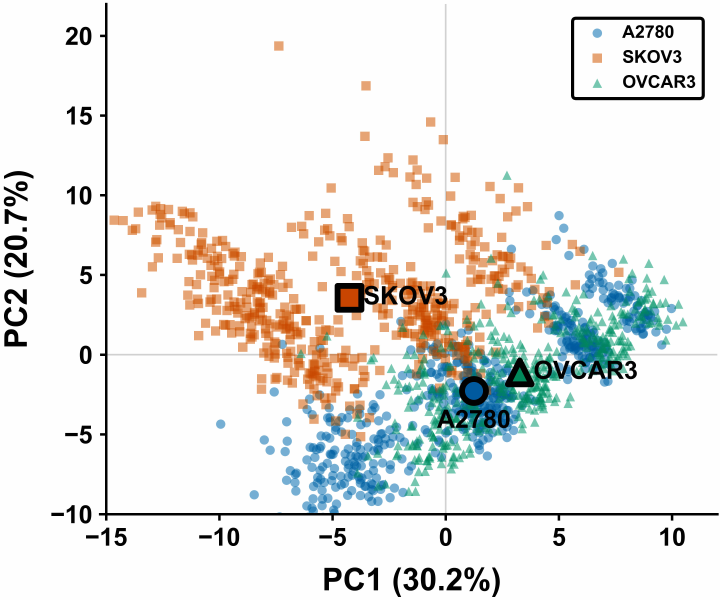}
    \put(-2,80){\large(b)}
    \end{overpic}
    \vspace{0.8em}
    %--------------------------------%
    \begin{overpic}[width=\linewidth]{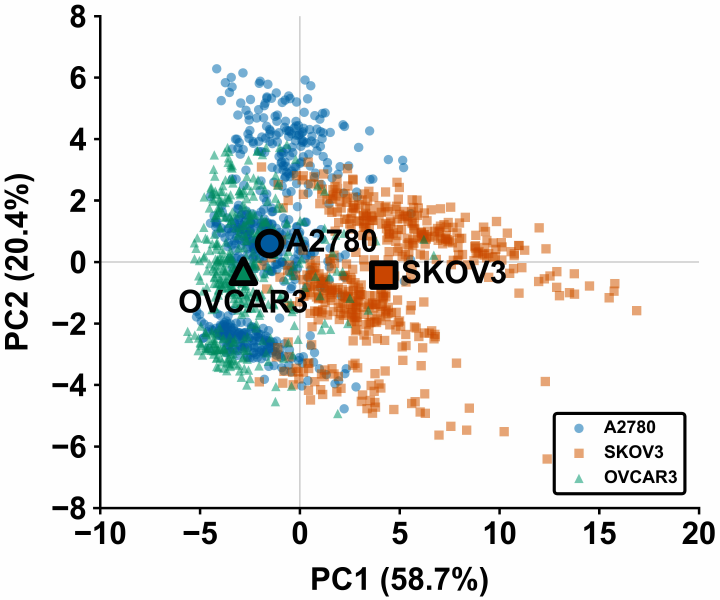}
    \put(-2,80){\large (c)}
    \end{overpic}
    %--------------------------------%    
    \end{minipage}
    %--------------------------------%
\caption{Cross-flow preservation of subtype-boundary organization in the refined Phenotype 2 feature space. (a) Cross-flow structural heatmap of the retained subset, showing ranking patterns of absolute Cliff's delta magnitudes for the three pairwise subtype comparisons. (b,c) PCA projections colored by cell type for the original full dynamic feature space and the retained 25-feature subset, respectively. The refined subset largely preserved the relative subtype geometry, with SKOV-3-related separations remaining more pronounced.}
\label{fig:HeatMap_CancerSubtypes}
\end{figure*}

To describe the specific inter-class topological relationships preserved within the refined subsets, cross-flow structural heatmaps were constructed (Figures~\ref{fig:HeatMap_Binary} and \ref{fig:HeatMap_CancerSubtypes}). For the binary classification task, interpretation centers on effect-size polarity. Each of the 20 optimized features strictly maintained a consistent directional relationship (i.e., uniformly $\text{Cancer} > \text{IOSE-80}$ or $\text{Cancer} < \text{IOSE-80}$) across the entire evaluated hydrodynamic spectrum (Figure~\ref{fig:HeatMap_Binary}a). This conserved polarity translates directly into the refined PCA projections, physically clarifying the centroid separation between the healthy epithelial baseline and the pooled malignant cohort (Figures~\ref{fig:HeatMap_Binary}b and \ref{fig:HeatMap_Binary}c). 

In the multi-class subtyping task, feature utility is dictated by the preservation of ordinal ranking among the three pairwise subtype boundaries. By ranking the absolute pairwise Cliff's delta magnitudes within each flow rate, the structural heatmap confirms that the refined descriptors recurrently prioritize specific subtype separations regardless of fluidic velocity (Figure~\ref{fig:HeatMap_CancerSubtypes}a). Separations involving the SKOV-3 lineage frequently dominate this ordinal hierarchy. This observation is corroborated by the corresponding PCA projections (Figures~\ref{fig:HeatMap_CancerSubtypes}b and \ref{fig:HeatMap_CancerSubtypes}c), where SKOV-3 exhibits pronounced spatial divergence, whereas the A2780 and OVCAR-3 clusters remain comparatively proximal. This topological geometry aligns rigorously with established mechanobiological profiles; SKOV-3 exhibits enhanced motility and unique mechanosensitive signaling, which collectively drive a divergent dynamic deformation response under extensional strain \cite{hallas2019ovarian, bileck2022inward}.

Mapping the physical identities of the retained features elucidates the biophysical strategies driving each diagnostic task. A fundamental subset of features was universally retained across both the binary and multi-class tasks, forming a consensus layer of stable mechanophenotypic readouts. This shared core is predominantly composed of intracellular optical-density metrics and transit kinematics, including localized grayscale states (e.g., $\mathrm{MeanGray}_{R}$, $\mathrm{MeanGray}_{X}$), contraction-to-release grayscale transitions (e.g., $d\mathrm{MeanGray}_{CR}$), and localized velocity profiles (e.g., $V$, $V_{\max}$, $V_{X}$, $d\mathrm{FV}_{RX}$). The recurrent selection of these variables indicates that macroscopic velocity perturbations and internal optical redistribution during microchannel transit capture fundamental, flow-invariant signatures of mechanical compliance.

Beyond this shared core, the refined subsets diverged to exploit task-specific biophysical characteristics. For the binary cancer-versus-healthy classification, the retained features are skewed toward inter-stage morphodynamic and kinematic transitions. This specific enrichment indicates that global malignant transformation is robustly characterized by transient cellular compliance and dynamic recovery rates as cells navigate spatial constrictions. Conversely, for multi-class subtyping, the discriminative architecture relies predominantly on stage-resolved static descriptors and localized intracellular heterogeneity. Selective retention of size, deformation, and subcellular optical density metrics indicates that distinguishing closely related cell lineages requires profiling baseline structural architecture and state-specific intracellular reorganization, not merely static macro-transition kinematics. Ultimately, these compositional mappings demonstrate that the stability-guided framework distills the raw high-dimensional space into distinct, physically meaningful mechanobiological signatures optimized for the targeted diagnostic granularity.

%========================================================%
%========================================================%

%========================================================%
%========================================================%

%---------------------------------------------------%
\subsection{Algorithm-Agnostic Validation and Sampling Robustness}
%---------------------------------------------------%

To establish that the optimized feature subsets capture intrinsic biophysical reality rather than overfitting to a specific algorithmic geometry, cross-model validation was executed. This evaluation utilized the pooled-flow dataset (integrating 10, 20, and 30~$\mu$L/h conditions) to simultaneously test model dependence and cross-condition stability. Four fundamentally distinct classifier architectures were employed: margin-based (Support Vector Machine, SVM), ensemble-based (Random Forest, RF), probabilistic linear (Logistic Regression, LR), and instance-based (K-Nearest Neighbors, KNN) algorithms \cite{nitta2018intelligent,ciaparrone2024label}.

\begin{figure*}[!t]
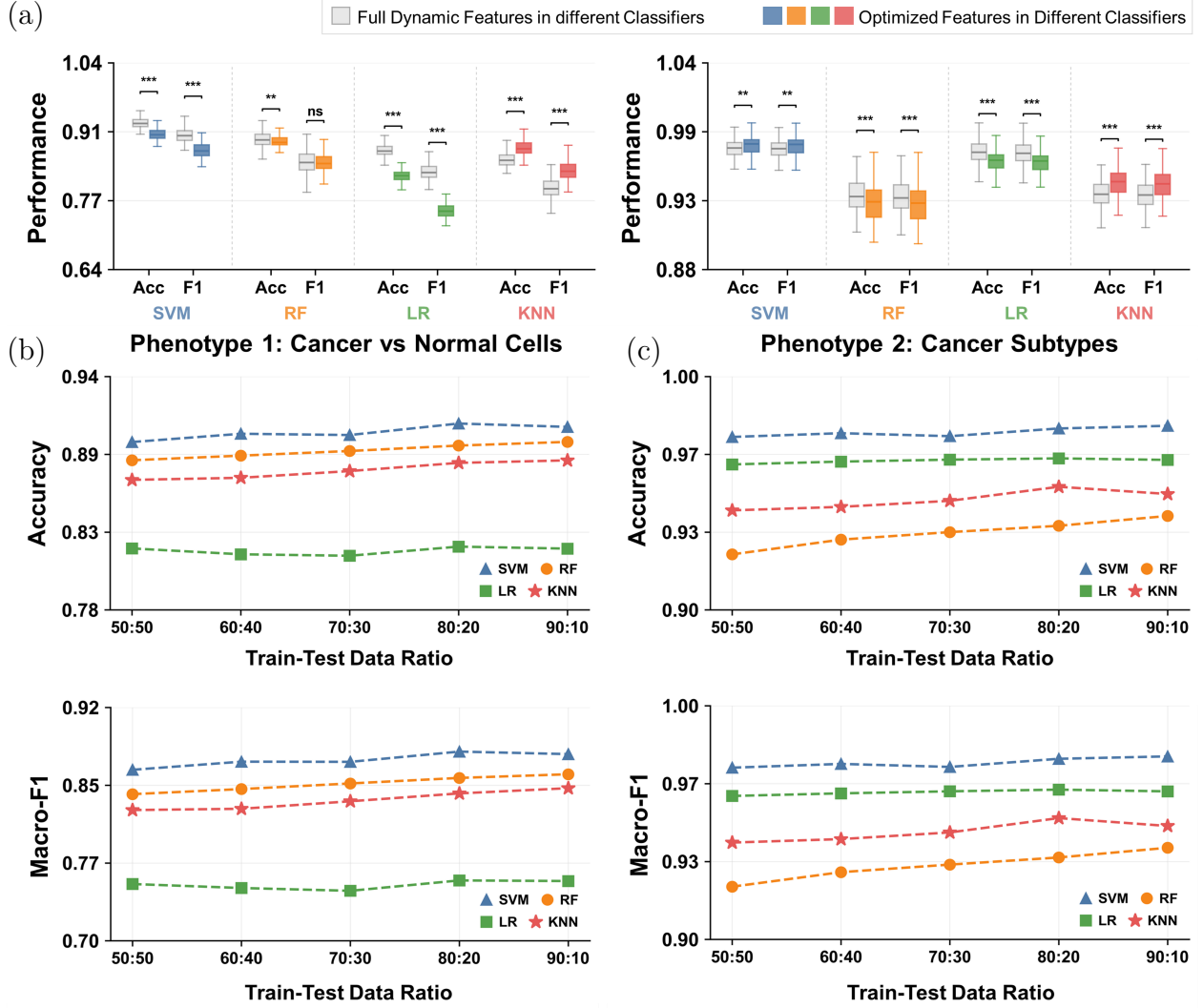

\centering
    %-------------------------------------------%
    \begin{overpic}[width=\linewidth]{Figure5bd.pdf}
    \put(0,25.5){\large (a)}
    \vspace{40pt}
    \end{overpic}
    %-------------------------------------------%
    \begin{overpic}[width=\linewidth]{Figure5ecfg.pdf}
    \put(0,55.5){\large (b)}
    \put(52,55.5){\large (c)}    
    \end{overpic}
    %-------------------------------------------%
\caption{Validation of cross-classifier and sampling-ratio robustness of the stability-guided retained feature subsets. (a) Classification performance of the original full dynamic feature spaces and the retained feature subsets across multiple classifiers (SVM, RF, LR, and KNN). Despite substantial dimensional reduction, the retained subsets maintained comparable Accuracy and Macro-F1 across models. Significance levels: \textsuperscript{ns}, $p \geq 0.05$; \textsuperscript{**}, $p < 0.01$; \textsuperscript{***}, $p < 0.001$. (b, c) Sampling-ratio robustness of the retained subsets under train-to-test split ratios from 50:50 to 90:10 at 20~$\mu$L/h. Line plots show mean Accuracy and Macro-F1 trends across SVM, RF, LR, and KNN classifiers using repeated stratified shuffle-split validation.}
\label{fig:figure4_panel2}
\end{figure*}

Across these diverse decision-making mechanisms, the optimized feature subsets maintained uniformly robust classification performance (Figure~\ref{fig:figure4_panel2}a). This cross-model stability was particularly critical for the binary classification task (Phenotype 1), which requires establishing a unified discriminative boundary against a highly heterogeneous malignant class (comprising pooled A2780, SKOV-3, and OVCAR-3 cells). Notably, even Logistic Regression—which enforces a highly restrictive linear decision boundary—retained strong diagnostic efficacy without exhibiting model-specific failure. In unrefined feature spaces, flow-induced perturbations often fold the data manifold, rendering class boundaries non-linear and requiring complex classifiers like SVMs or RFs to resolve. By systematically rejecting flow-coupled variables, the refinement process flattens this manifold, allowing straightforward hyperplanes to achieve comparable discrimination. For the multi-class subtyping task (Phenotype 2), the 25-feature subset similarly supported high performance across all architectures, confirming the algorithm-independent validity of the retained variables.

Beyond classifier architecture, practical clinical and microfluidic applications are frequently constrained by limited or imbalanced cell populations. To assess resilience against data scarcity, the sampling robustness of the optimized subsets was evaluated at the median flow rate of 20~$\mu$L/h using progressively restricted training-to-testing split ratios ranging from 50:50 to 90:10 (Figures \ref{fig:figure4_panel2}b and \ref{fig:figure4_panel2}c). Across this sampling spectrum, both accuracy and Macro-$F_1$ metrics exhibited stability. Maintaining diagnostic fidelity under a balanced 50:50 data split indicates that the retained features provide high information density; they do not require massive, data-hungry validation sets to establish reliable classification boundaries. This insensitivity to data-subsampling regimes demonstrates the practical generalization of the optimized subsets, addressing a common limitation in translational microfluidics, where target cell populations are frequently scarce or imbalanced \cite{sivakumar2024trade}.

%========================================================%
%========================================================%

%---------------------------------------------------%
\subsection{Diagnostic Validation of the Refined Feature Subsets}
%---------------------------------------------------%

\begin{figure*}[!ht]
\centering
% Figure 3a-b
\begin{overpic}[width=0.455\textwidth]{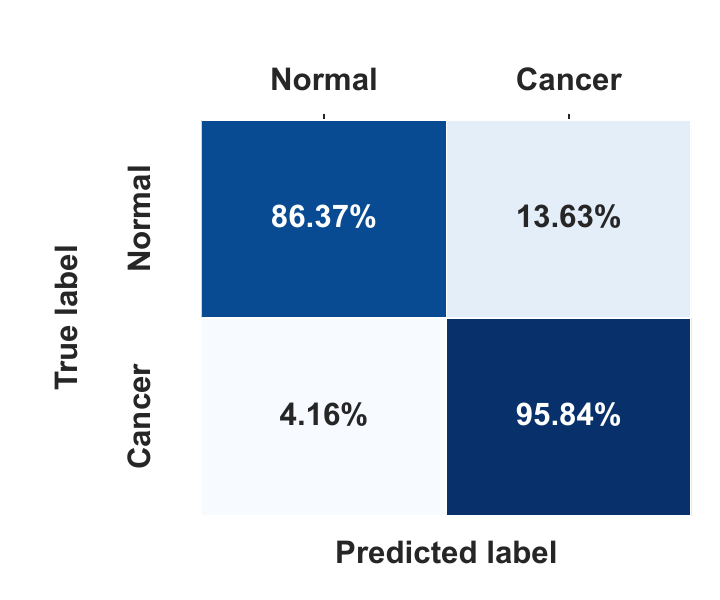}
\put(0,72){\large{(a)}}
\end{overpic}
\hfill
\begin{overpic}[width=0.455\textwidth]{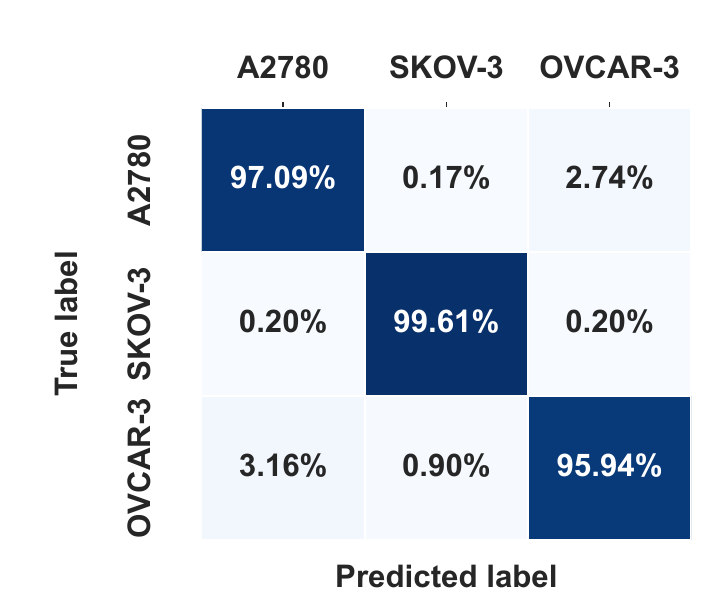}
\put(0,72){\large{(b)}}
\end{overpic}
%--------------------------------------%
\vspace{0.75em}
\centering
% Figure 3c-d
\begin{overpic}[width=0.455\textwidth]{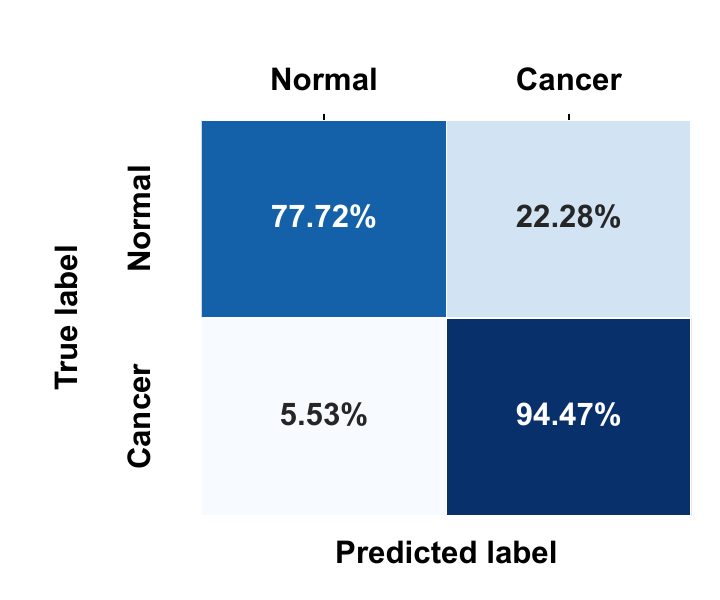}
\put(0,72){\large{(c)}}
\end{overpic}
\hfill
\begin{overpic}[width=0.455\textwidth]{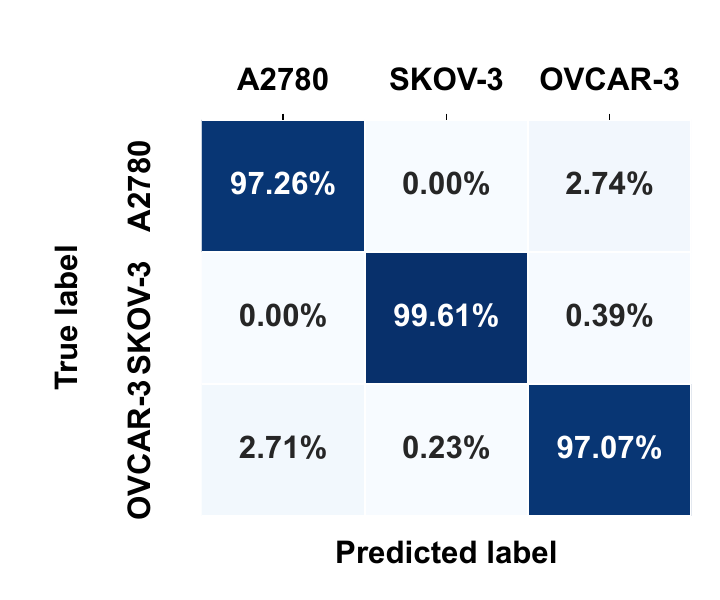}
\put(0,72){\large{(d)}}
\end{overpic}
%--------------------------------------%
\vspace{0.75em}
\caption{Classification performance of the original and refined feature spaces. (a,b) Confusion matrices for Phenotype 1 and Phenotype 2 classification using the original full dynamic feature spaces. (c,d) Confusion matrices for Phenotype 1 and Phenotype 2 classification using the retained 20-feature and 25-feature subsets after stability-guided refinement, respectively. The retained subsets maintained the principal diagnostic separations observed in the original high-dimensional feature spaces.}
\label{fig:PCA_FlowRate_confusion}
\end{figure*}

The analytical utility of the refined feature subsets was validated against the two predefined diagnostic objectives: binary cancer-versus-healthy discrimination (Phenotype 1) and multi-class malignant subtyping (Phenotype 2) (Figure~\ref{fig:PCA_FlowRate_confusion}a--d). Following stability-guided dimensional compression, the optimized 20- and 25-feature subsets successfully encapsulated the primary diagnostic boundaries of the global 93-dimensional feature space.

For the binary classification task (Phenotype 1), the optimized 20-feature subset prioritized malignant sensitivity, maintaining a cancer-cell recall of 94.47\% (compared with 95.84\% in the unrefined space). Healthy-cell recall underwent a moderate shift from 86.37\% to 77.72\%, corresponding to a slight compression in overall accuracy (93.37\% to 90.10\%) and Macro-$F_1$ score (91.35\% to 86.88\%). This minor performance offset represents an acceptable regularization trade-off in high-dimensional analysis. The unrefined feature space likely overfitted to transient batch effects or fluidic artifacts to artificially inflate healthy-cell recall. By enforcing strict cross-flow stability, the 20-feature subset provides a generalized, mathematically conservative diagnostic boundary that reliably detects malignancy while mitigating artifact-driven overfitting.

Conversely, for the multi-class task (Phenotype 2), the optimized 25-feature subset exhibited essentially unchanged subtype-level performance. Class-wise recalls were high, reaching 97.26\%, 99.61\%, and 97.07\% for A2780, SKOV-3, and OVCAR-3, respectively. Aggregate metrics remained statistically comparable to the full feature space, with overall accuracy shifting from 97.59\% to 97.98\% and the Macro-$F_1$ score transitioning from 97.54\% to 97.94\%. 

Examination of the confusion matrices provides biological context for the refinement process. Residual misclassification errors strictly localized between the A2780 and OVCAR-3 populations, whereas the SKOV-3 lineage remained almost entirely isolated. This error distribution corroborates the structural topology observed in the preceding PCA and ranking-order heatmaps (Figures~\ref{fig:HeatMap_CancerSubtypes}a--c). Ovarian cancer lines such as A2780 and OVCAR-3 often share similar macroscopic deformability profiles under continuous flow. Their residual classification overlap reflects an authentic biophysical similarity rather than a failure of the feature-selection protocol. By preserving this expected biological overlap while maintaining aggregate accuracy, the refined feature representation proves to be a biologically faithful representation of the cellular populations, successfully distilling critical phenotypic resolution without relying on fluidic noise.

%========================================================%
%========================================================%

%---------------------------------------------------%
\section{Conclusions and Perspectives}
%---------------------------------------------------%

In this study, we developed an analytical framework for label-free cellular mechanophenotyping that systematically isolates intrinsic biological signatures from hydrodynamic artifacts. By integrating high-speed bright-field imaging within a hyperbolic microchannel with a stability-guided feature-selection protocol, we addressed the critical confounding effects of flow-rate variations inherent to high-dimensional dynamic feature spaces. 

Our findings demonstrate that while continuous cellular deformation trajectories contain rich phenotypic information, unrefined high-dimensional representations are heavily confounded by operational fluidic conditions. By enforcing cross-flow structural consistency and statistical persistence via non-parametric effect-size evaluations, the proposed refinement strategy successfully decoupled these fluidic artifacts. Variance partitioning mathematically confirmed this decoupling, demonstrating that the primary principal components of the refined subsets transitioned from being artifactually flow-dominated to strictly class-associated. Furthermore, compositional mapping of the retained subsets revealed that distinct diagnostic granularities exploit different biophysical strategies: macroscopic healthy-versus-malignant discrimination (Phenotype 1) relies predominantly on global morphodynamic and kinematic transitions, whereas fine-grained malignant subtyping (Phenotype 2) necessitates the high-resolution profiling of localized intracellular optical densities.

The analytical robustness of the optimized 20- and 25-feature subsets was validated across diverse machine-learning architectures (SVM, RF, LR, and KNN) and under restricted training-data sampling ratios. The consistent classification performance across these disparate conditions confirms that the retained descriptors capture fundamental biophysical reality rather than algorithm-specific geometric boundaries. By preserving characteristic inter-class topological structures while reducing the feature space by approximately 75\%, the framework inherently mitigates the risk of artifact-driven overfitting typically associated with high-throughput single-cell screening.

Looking forward, the capability to extract flow-invariant mechanobiological descriptors establishes a critical foundation for translating microfluidic phenotyping to complex clinical settings. Because target cell populations in physiological fluids are frequently scarce, heterogeneous, and subjected to variable micro-environmental transport conditions, relying on cross-condition validated feature subsets is essential for robust diagnostic profiling. Future efforts will focus on evaluating the generalizability of this framework across highly heterogeneous patient-derived samples and integrating the stability-guided algorithms with real-time, continuous-flow cell sorting mechanisms. Ultimately, this study indicates that shifting the analytical focus from maximizing feature dimensionality toward optimizing cross-condition biophysical stability provides a more reliable and physically interpretable pathway for label-free cellular phenotyping.

%---------------------------------------------------%
\section{Declaration of Competing Interest}
%---------------------------------------------------%
The authors declare that they have no known competing financial interests or personal relationships that could have appeared to influence the work reported in this paper.

%---------------------------------------------------%
\section*{Author Contributions}
%---------------------------------------------------%
\textbf{Ms. Hong-Fei LI}: Methodology, Software, Formal Analysis, Data Curation, Investigation, Writing – Original Draft. \\
\textbf{Mr. Xi-Lin GAO}: Validation, Visualization, Writing – Original Draft. \\
\textbf{Ms. Yi-Juan XIANG}: Methodology, Writing – Review \& Editing. \\
\textbf{Mr. Shu-Song HUANG}: Software, Validation. \\
\textbf{Mr. Yi-Lin WANG}: Validation, Writing – Review \& Editing. \\
\textbf{Assoc. Prof. Dr. Chun-Dong XUE}: Resources. \\
\textbf{Prof. Dr. Zhuo YANG}: Resources, Supervision, Funding Acquisition. \\
\textbf{Assoc. Prof. Dr. Yong-Jiang LI}: Supervision, Writing – Review \& Editing. \\
\textbf{Prof. Dr. Xu-Qu HU}: Conceptualization, Methodology, Funding Acquisition, Project Administration, Resources, Supervision, Writing – Review \& Editing.

%---------------------------------------------------%
\section{Acknowledgements}
%---------------------------------------------------%
This research was supported by the Fundamental Research Funds for the Central Universities (DUT24RC(3)106), the Dalian Municipal Guiding Program for the Life and Health Sector (2025ZDJH01PT040), the Science and Technology Plan (joint plan) of Liaoning Provincial Department of Science and Technology (2024JH2/102600176) and the National Natural Science Foundation of China (12472306).

%\newpage
%========================================================%
\printbibliography
% \bibliography{2026DynaFeatureOPT}
%========================================================%
%\newpage

% \clearpage
% \appendix
% \section*{Supporting Information}

% \clearpage
\subsection*{Supplementary Tables}

% Table S1: already contains longtable, caption, and label
\begingroup
\scriptsize
\setlength{\tabcolsep}{2.5pt}
\renewcommand{\arraystretch}{1.08}
\sloppy
\begin{longtable}{p{2.20cm}p{1.80cm}p{1.60cm}p{6.20cm}}
\caption{Mathematical equations of the Full  feature space.} \label{tab:S1_93D_features} \\
\toprule
Feature & Category & Stage & Equation \\

\midrule
\endfirsthead
\caption[]{Mathematical equations of the 93-dimensional dynamic feature set.} \\
\toprule
Feature & Category & Stage & Equation \\
\midrule
\endhead
\midrule
\multicolumn{4}{r}{Continued on next page} \\
\midrule
\endfoot
\bottomrule
\endlastfoot
\texttt{V} & Kin. & E-C-R-X & \(V=N^{-1}\sum_{i=1}^{N}V_i\) \\
\texttt{FV} & Kin. & E-C-R-X & \(FV=\left[(N-1)^{-1}\sum_{i=1}^{N}(V_i-V)^2\right]^{1/2}\) \\
\texttt{V\_max} & Kin. & E-C-R-X & \(V_{\max}=\max_{1\le i\le N}V_i\) \\
\texttt{V\_min} & Kin. & E-C-R-X & \(V_{\min}=\min_{1\le i\le N}V_i\) \\
\texttt{A} & Kin. & E-C-R-X & \(A=N^{-1}\sum_{i=1}^{N}A_i\) \\
\texttt{FA} & Kin. & E-C-R-X & \(FA=\left[(N-1)^{-1}\sum_{i=1}^{N}(A_i-A)^2\right]^{1/2}\) \\
\texttt{A\_max} & Kin. & E-C-R-X & \(A_{\max}=\max_{1\le i\le N}A_i\) \\
\texttt{A\_min} & Kin. & E-C-R-X & \(A_{\min}=\min_{1\le i\le N}A_i\) \\
\texttt{MeanGray} & Gray. & E-C-R-X & \(\mathrm{MeanGray}=N^{-1}\sum_{i=1}^{N}\mathrm{MeanGray}_i\) \\
\texttt{VarGray} & Gray. & E-C-R-X & \(\mathrm{VarGray}=N^{-1}\sum_{i=1}^{N}\mathrm{VarGray}_i\) \\
\texttt{CVGray} & Gray. & E-C-R-X & \(\mathrm{CVGray}=N^{-1}\sum_{i=1}^{N}\mathrm{CVGray}_i\) \\
\texttt{Roundness} & Morph. & E-C-R-X & \(\mathrm{Roundness}=N^{-1}\sum_{i=1}^{N}\mathrm{Roundness}_i\) \\
\texttt{AspectRatio} & Morph. & E-C-R-X & \(\mathrm{AspectRatio}=N^{-1}\sum_{i=1}^{N}\mathrm{AspectRatio}_i\) \\
\texttt{Taylor\_ECRX} & Morph. & E-C-R-X & \(\mathrm{Taylor}_{\mathrm{ECRX}}=N^{-1}\sum_{i=1}^{N}\mathrm{Taylor}_i\) \\
\texttt{V\_E} & Kin. & E & \(V_{E}=N_{E}^{-1}\sum_{i\in E}V_i\) \\
\texttt{FV\_E} & Kin. & E & \(FV_{E}=\left[(N_{E}-1)^{-1}\sum_{i\in E}(V_i-V_{E})^2\right]^{1/2}\) \\
\texttt{A\_E} & Kin. & E & \(A_{E}=N_{E}^{-1}\sum_{i\in E}A_i\) \\
\texttt{FA\_E} & Kin. & E & \(FA_{E}=\left[(N_{E}-1)^{-1}\sum_{i\in E}(A_i-A_{E})^2\right]^{1/2}\) \\
\texttt{MeanGray\_E} & Gray. & E & \(\mathrm{MeanGray}_{E}=N_{E}^{-1}\sum_{i\in E}\mathrm{MeanGray}_i\) \\
\texttt{VarGray\_E} & Gray. & E & \(\mathrm{VarGray}_{E}=N_{E}^{-1}\sum_{i\in E}\mathrm{VarGray}_i\) \\
\texttt{CVGray\_E} & Gray. & E & \(\mathrm{CVGray}_{E}=N_{E}^{-1}\sum_{i\in E}\mathrm{CVGray}_i\) \\
\texttt{Roundness\_E} & Morph. & E & \(\mathrm{Roundness}_{E}=N_{E}^{-1}\sum_{i\in E}\mathrm{Roundness}_i\) \\
\texttt{AspectRatio\_E} & Morph. & E & \(\mathrm{AspectRatio}_{E}=N_{E}^{-1}\sum_{i\in E}\mathrm{AspectRatio}_i\) \\
\texttt{Taylor\_E} & Morph. & E & \(\mathrm{Taylor}_{E}=N_{E}^{-1}\sum_{i\in E}\mathrm{Taylor}_i\) \\
\texttt{Deq\_E} & Morph. & E & \(D_{\mathrm{eq},E}=N_{E}^{-1}\sum_{i\in E}D_{\mathrm{eq},i}\) \\
\texttt{V\_C} & Kin. & C & \(V_{C}=N_{C}^{-1}\sum_{i\in C}V_i\) \\
\texttt{FV\_C} & Kin. & C & \(FV_{C}=\left[(N_{C}-1)^{-1}\sum_{i\in C}(V_i-V_{C})^2\right]^{1/2}\) \\
\texttt{A\_C} & Kin. & C & \(A_{C}=N_{C}^{-1}\sum_{i\in C}A_i\) \\
\texttt{FA\_C} & Kin. & C & \(FA_{C}=\left[(N_{C}-1)^{-1}\sum_{i\in C}(A_i-A_{C})^2\right]^{1/2}\) \\
\texttt{MeanGray\_C} & Gray. & C & \(\mathrm{MeanGray}_{C}=N_{C}^{-1}\sum_{i\in C}\mathrm{MeanGray}_i\) \\
\texttt{VarGray\_C} & Gray. & C & \(\mathrm{VarGray}_{C}=N_{C}^{-1}\sum_{i\in C}\mathrm{VarGray}_i\) \\
\texttt{CVGray\_C} & Gray. & C & \(\mathrm{CVGray}_{C}=N_{C}^{-1}\sum_{i\in C}\mathrm{CVGray}_i\) \\
\texttt{Roundness\_C} & Morph. & C & \(\mathrm{Roundness}_{C}=N_{C}^{-1}\sum_{i\in C}\mathrm{Roundness}_i\) \\
\texttt{AspectRatio\_C} & Morph. & C & \(\mathrm{AspectRatio}_{C}=N_{C}^{-1}\sum_{i\in C}\mathrm{AspectRatio}_i\) \\
\texttt{Taylor\_C} & Morph. & C & \(\mathrm{Taylor}_{C}=N_{C}^{-1}\sum_{i\in C}\mathrm{Taylor}_i\) \\
\texttt{Deq\_C} & Morph. & C & \(D_{\mathrm{eq},C}=N_{C}^{-1}\sum_{i\in C}D_{\mathrm{eq},i}\) \\
\texttt{V\_R} & Kin. & R & \(V_{R}=N_{R}^{-1}\sum_{i\in R}V_i\) \\
\texttt{FV\_R} & Kin. & R & \(FV_{R}=\left[(N_{R}-1)^{-1}\sum_{i\in R}(V_i-V_{R})^2\right]^{1/2}\) \\
\texttt{A\_R} & Kin. & R & \(A_{R}=N_{R}^{-1}\sum_{i\in R}A_i\) \\
\texttt{FA\_R} & Kin. & R & \(FA_{R}=\left[(N_{R}-1)^{-1}\sum_{i\in R}(A_i-A_{R})^2\right]^{1/2}\) \\
\texttt{MeanGray\_R} & Gray. & R & \(\mathrm{MeanGray}_{R}=N_{R}^{-1}\sum_{i\in R}\mathrm{MeanGray}_i\) \\
\texttt{VarGray\_R} & Gray. & R & \(\mathrm{VarGray}_{R}=N_{R}^{-1}\sum_{i\in R}\mathrm{VarGray}_i\) \\
\texttt{CVGray\_R} & Gray. & R & \(\mathrm{CVGray}_{R}=N_{R}^{-1}\sum_{i\in R}\mathrm{CVGray}_i\) \\
\texttt{Roundness\_R} & Morph. & R & \(\mathrm{Roundness}_{R}=N_{R}^{-1}\sum_{i\in R}\mathrm{Roundness}_i\) \\
\texttt{AspectRatio\_R} & Morph. & R & \(\mathrm{AspectRatio}_{R}=N_{R}^{-1}\sum_{i\in R}\mathrm{AspectRatio}_i\) \\
\texttt{Taylor\_R} & Morph. & R & \(\mathrm{Taylor}_{R}=N_{R}^{-1}\sum_{i\in R}\mathrm{Taylor}_i\) \\
\texttt{Deq\_R} & Morph. & R & \(D_{\mathrm{eq},R}=N_{R}^{-1}\sum_{i\in R}D_{\mathrm{eq},i}\) \\
\texttt{V\_X} & Kin. & X & \(V_{X}=N_{X}^{-1}\sum_{i\in X}V_i\) \\
\texttt{FV\_X} & Kin. & X & \(FV_{X}=\left[(N_{X}-1)^{-1}\sum_{i\in X}(V_i-V_{X})^2\right]^{1/2}\) \\
\texttt{A\_X} & Kin. & X & \(A_{X}=N_{X}^{-1}\sum_{i\in X}A_i\) \\
\texttt{FA\_X} & Kin. & X & \(FA_{X}=\left[(N_{X}-1)^{-1}\sum_{i\in X}(A_i-A_{X})^2\right]^{1/2}\) \\
\texttt{MeanGray\_X} & Gray. & X & \(\mathrm{MeanGray}_{X}=N_{X}^{-1}\sum_{i\in X}\mathrm{MeanGray}_i\) \\
\texttt{VarGray\_X} & Gray. & X & \(\mathrm{VarGray}_{X}=N_{X}^{-1}\sum_{i\in X}\mathrm{VarGray}_i\) \\
\texttt{CVGray\_X} & Gray. & X & \(\mathrm{CVGray}_{X}=N_{X}^{-1}\sum_{i\in X}\mathrm{CVGray}_i\) \\
\texttt{Roundness\_X} & Morph. & X & \(\mathrm{Roundness}_{X}=N_{X}^{-1}\sum_{i\in X}\mathrm{Roundness}_i\) \\
\texttt{AspectRatio\_X} & Morph. & X & \(\mathrm{AspectRatio}_{X}=N_{X}^{-1}\sum_{i\in X}\mathrm{AspectRatio}_i\) \\
\texttt{Taylor\_X} & Morph. & X & \(\mathrm{Taylor}_{X}=N_{X}^{-1}\sum_{i\in X}\mathrm{Taylor}_i\) \\
\texttt{Deq\_X} & Morph. & X & \(D_{\mathrm{eq},X}=N_{X}^{-1}\sum_{i\in X}D_{\mathrm{eq},i}\) \\
\texttt{dV\_EC} & Kin. & E-C & \(\Delta V_{E-C}=V_{C}-V_{E}\) \\
\texttt{dFV\_EC} & Kin. & E-C & \(\Delta FV_{E-C}=FV_{C}-FV_{E}\) \\
\texttt{dA\_EC} & Kin. & E-C & \(\Delta A_{E-C}=A_{C}-A_{E}\) \\
\texttt{dFA\_EC} & Kin. & E-C & \(\Delta FA_{E-C}=FA_{C}-FA_{E}\) \\
\texttt{dMeanGray\_EC} & Gray. & E-C & \(\Delta \mathrm{MeanGray}_{E-C}=\mathrm{MeanGray}_{C}-\mathrm{MeanGray}_{E}\) \\
\texttt{dVarGray\_EC} & Gray. & E-C & \(\Delta \mathrm{VarGray}_{E-C}=\mathrm{VarGray}_{C}-\mathrm{VarGray}_{E}\) \\
\texttt{dCVGray\_EC} & Gray. & E-C & \(\Delta \mathrm{CVGray}_{E-C}=\mathrm{CVGray}_{C}-\mathrm{CVGray}_{E}\) \\
\texttt{dRoundness\_EC} & Morph. & E-C & \(\Delta \mathrm{Roundness}_{E-C}=\mathrm{Roundness}_{C}-\mathrm{Roundness}_{E}\) \\
\texttt{dAspectRatio\_EC} & Morph. & E-C & \(\Delta \mathrm{AR}_{E-C}=\mathrm{AR}_{C}-\mathrm{AR}_{E}\) \\
\texttt{dTaylor\_EC} & Morph. & E-C & \(\Delta \mathrm{Taylor}_{E-C}=\mathrm{Taylor}_{C}-\mathrm{Taylor}_{E}\) \\
\texttt{dV\_CR} & Kin. & C-R & \(\Delta V_{C-R}=V_{R}-V_{C}\) \\
\texttt{dFV\_CR} & Kin. & C-R & \(\Delta FV_{C-R}=FV_{R}-FV_{C}\) \\
\texttt{dA\_CR} & Kin. & C-R & \(\Delta A_{C-R}=A_{R}-A_{C}\) \\
\texttt{dFA\_CR} & Kin. & C-R & \(\Delta FA_{C-R}=FA_{R}-FA_{C}\) \\
\texttt{dMeanGray\_CR} & Gray. & C-R & \(\Delta \mathrm{MeanGray}_{C-R}=\mathrm{MeanGray}_{R}-\mathrm{MeanGray}_{C}\) \\
\texttt{dVarGray\_CR} & Gray. & C-R & \(\Delta \mathrm{VarGray}_{C-R}=\mathrm{VarGray}_{R}-\mathrm{VarGray}_{C}\) \\
\texttt{dCVGray\_CR} & Gray. & C-R & \(\Delta \mathrm{CVGray}_{C-R}=\mathrm{CVGray}_{R}-\mathrm{CVGray}_{C}\) \\
\texttt{dRoundness\_CR} & Morph. & C-R & \(\Delta \mathrm{Roundness}_{C-R}=\mathrm{Roundness}_{R}-\mathrm{Roundness}_{C}\) \\
\texttt{dAspectRatio\_CR} & Morph. & C-R & \(\Delta \mathrm{AspectRatio}_{C-R}=\mathrm{AR.}_{R}-\mathrm{AR.}_{C}\) \\
\texttt{dTaylor\_CR} & Morph. & C-R & \(\Delta \mathrm{Taylor}_{C-R}=\mathrm{Taylor}_{R}-\mathrm{Taylor}_{C}\) \\
\texttt{dV\_RX} & Kin. & R-X & \(\Delta V_{R-X}=V_{X}-V_{R}\) \\
\texttt{dFV\_RX} & Kin. & R-X & \(\Delta FV_{R-X}=FV_{X}-FV_{R}\) \\
\texttt{dA\_RX} & Kin. & R-X & \(\Delta A_{R-X}=A_{X}-A_{R}\) \\
\texttt{dFA\_RX} & Kin. & R-X & \(\Delta FA_{R-X}=FA_{X}-FA_{R}\) \\
\texttt{dMeanGray\_RX} & Gray. & R-X & \(\Delta \mathrm{MeanGray}_{R-X}=\mathrm{MeanGray}_{X}-\mathrm{MeanGray}_{R}\) \\
\texttt{dVarGray\_RX} & Gray. & R-X & \(\Delta \mathrm{VarGray}_{R-X}=\mathrm{VarGray}_{X}-\mathrm{VarGray}_{R}\) \\
\texttt{dCVGray\_RX} & Gray. & R-X & \(\Delta \mathrm{CVGray}_{R-X}=\mathrm{CVGray}_{X}-\mathrm{CVGray}_{R}\) \\
\texttt{dRoundness\_RX} & Morph. & R-X & \(\Delta \mathrm{Roundness}_{R-X}=\mathrm{Roundness}_{X}-\mathrm{Roundness}_{R}\) \\
\texttt{dAspectRatio\_RX} & Morph. & R-X & \(\Delta \mathrm{AspectRatio}_{R-X}=\mathrm{AR.}_{X}-\mathrm{AR.}_{R}\) \\
\texttt{dTaylor\_RX} & Morph. & R-X & \(\Delta \mathrm{Taylor}_{R-X}=\mathrm{Taylor}_{X}-\mathrm{Taylor}_{R}\) \\
\texttt{Deq\_ratio\_EX} & Morph. & E-X & \(D_{\mathrm{eq,ratio},E-X}=(D_{\mathrm{eq},E}-D_{\mathrm{eq},X})/(D_{\mathrm{eq},E}+\epsilon)\) \\
\texttt{t\_decay50\_R} & Temporal/Kin. & R & \(t_{\mathrm{decay50},R}=t_{50,R}-t_{\mathrm{peak},R}\), with \(V(t_{50,R})=0.5V_{\mathrm{peak},R}\) \\
\texttt{plateau\_dur\_R} & Temporal/Kin. & R & \(t_{\mathrm{plateau},R}=\max_k(t^{(k)}_{\mathrm{end}}-t^{(k)}_{\mathrm{start}})\), where \(|dV/dt|\le0.05\max_{t\in R}|dV/dt|\) \\
\texttt{tau\_decay\_R} & Temporal/Kin. & R & \(V(t)=V_{\infty,R}+(V_{\mathrm{peak},R}-V_{\infty,R})\exp[-(t-t_{\mathrm{peak},R})/\tau_{\mathrm{decay},R}]\) \\
%------------------------------------------------------------%
\texttt{tau\_recovery\_R} & Temporal/Kin. & R & \(V_{\mathrm{end},R}-V(t)=(V_{\mathrm{end},R}-V_{\min,R})\exp[-(t-t_{\min,R})/\tau_{\mathrm{recovery},R}]\) \\
% $\(V_{\mathrm{end},R}-V(t)=\left(V_{\mathrm{end},R}-V_{\min,R}\right)e^{-(t-t_{\min,R})/\tau_{\mathrm{recovery},R}}\)$ \\
%------------------------------------------------------------%
\end{longtable}
\endgroup
%------------------------------------------------------------%
%------------------------------------------------------------%

% \noindent \textbf{Notes:}
% \begin{itemize}
%     \item \textbf{Stage notation.} E, C, R, and X denote the Entry, Compression, Release, and Exit regions, respectively. Stage assignments were based on the cell-center position \(c_x\): E(300--330) \(\mu\)m, C(100--300) \(\mu\)m, R(-300 -- -100) \(\mu\)m, X(-330 -- -300) \(\mu\)m. \(N_S\) is the number of valid frames in stage \(S\); \(N\) is the total number of valid frames across the full E--C--R--X trajectory.
% \end{itemize}

% Table S4
\begin{table}[htbp]
\centering
\caption{Structural summary of the descriptor-specific 77-dimensional feature subspaces used for Figure 4. The Roundness, Taylor, and Aspect-ratio subspaces were constructed using the same structural template, whereas the full dynamic feature set was included as the 93-dimensional reference feature set for comparison.}
\label{Tab:77features}
\resizebox{\textwidth}{!}{\small
\begin{tabular}{llllp{5.6cm}p{5.4cm}}
\hline
Subspace & Program label & Descriptor basis & Dimension & Structural organization & Purpose in Figure 4 \\
\hline
Roundness subspace & Group1\_Roundness & Roundness & 77 & Stage-wise statistics, inter-stage differences, grayscale descriptors, and kinematic descriptors & Descriptor-specific comparison and representative binary-task stability screening \\
Taylor subspace & Group2\_Taylor & Taylor parameter & 77 & Stage-wise statistics, inter-stage differences, grayscale descriptors, and kinematic descriptors & Descriptor-specific comparison and representative subtype-task stability screening \\
Aspect-ratio subspace & Group3\_AspectRatio & Aspect ratio & 77 & Stage-wise statistics, inter-stage differences, grayscale descriptors, and kinematic descriptors & Descriptor-specific comparison \\
Full dynamic feature set & All\_93 & Mixed descriptors & 93 & Full morphodynamic, grayscale, and kinematic feature architecture & Reference feature set for comparison \\
\hline
\end{tabular}
}
\end{table}

% Table S6a
\begin{table}[htbp]
\centering
\caption{\textbf{Supplementary Table S6a.} Stability-guided feature subset selected for cancer-versus-normal classification. The relative stable effect size denotes the mean absolute effect size above the task-specific minimum-effect threshold used in the stability-selection procedure. Structural indicates whether the feature satisfied the task-specific structural-consistency criterion, and CI indicates whether the bootstrap confidence interval excluded zero across all flow rates.}
\label{tab:S6a}
\small
\begin{tabular}{lclcc}
\hline
Feature & Family & Relative stable effect size & Structural & CI \\
\hline
dMeanGray\_CR & Grayscale & 0.5191 & yes & yes \\
dMeanGray\_RX & Grayscale & 0.4157 & yes & yes \\
dCVGray\_CR & Grayscale & 0.3582 & yes & yes \\
FV\_R & Kinematics & 0.3360 & no & yes \\
dFV\_EC & Kinematics & 0.3146 & no & yes \\
V\_max & Kinematics & 0.3143 & no & yes \\
dFV\_RX & Kinematics & 0.3127 & no & yes \\
FV\_C & Kinematics & 0.3068 & no & yes \\
FV & Kinematics & 0.2949 & no & yes \\
dV\_RX & Kinematics & 0.2755 & no & yes \\
V & Kinematics & 0.2706 & yes & yes \\
V\_E & Kinematics & 0.2560 & yes & yes \\
V\_C & Kinematics & 0.2492 & yes & no \\
V\_X & Kinematics & 0.2411 & yes & no \\
MeanGray\_X & Grayscale & 0.2377 & yes & yes \\
dVarGray\_CR & Grayscale & 0.2268 & yes & yes \\
dA\_RX & Kinematics & 0.1987 & no & yes \\
MeanGray\_R & Grayscale & 0.1721 & yes & yes \\
dA\_CR & Kinematics & 0.1563 & yes & yes \\
Roundness & Morphology & 0.1474 & no & yes \\
\hline
\end{tabular}

\end{table}

% Table S6b
\begin{table}[htbp]
\centering
\caption{\textbf{Supplementary Table S6b.} Stability-guided feature subset selected for cancer-subtype classification. The relative stable effect size denotes the mean absolute effect size above the task-specific minimum-effect threshold used in the stability-selection procedure. Structural indicates whether the feature satisfied the task-specific structural-consistency criterion, and CI indicates whether the bootstrap confidence interval excluded zero across all flow rates.}
\label{tab:S6b}
\small
\begin{tabular}{l c l c c}
\hline
Feature & Family & Relative stable effect size & Structural & CI \\
\hline
CVGray\_X & Grayscale & 0.6852 & yes & yes \\
CVGray\_R & Grayscale & 0.6751 & yes & yes \\
CVGray & Grayscale & 0.6711 & yes & yes \\
CVGray\_C & Grayscale & 0.6416 & yes & yes \\
CVGray\_E & Grayscale & 0.6338 & yes & yes \\
Deq\_X & Morphology & 0.6293 & yes & yes \\
VarGray\_X & Grayscale & 0.6243 & yes & yes \\
VarGray\_R & Grayscale & 0.6189 & no & yes \\
VarGray & Grayscale & 0.6127 & no & yes \\
Deq\_R & Morphology & 0.6089 & yes & yes \\
VarGray\_C & Grayscale & 0.5915 & no & yes \\
VarGray\_E & Grayscale & 0.5910 & yes & yes \\
Deq\_E & Morphology & 0.5683 & yes & yes \\
Deq\_C & Morphology & 0.5459 & yes & yes \\
dMeanGray\_CR & Grayscale & 0.5318 & yes & yes \\
V\_max & Kinematics & 0.4959 & no & yes \\
MeanGray\_X & Grayscale & 0.4638 & no & yes \\
V\_X & Kinematics & 0.4614 & no & yes \\
MeanGray\_R & Grayscale & 0.4501 & no & yes \\
MeanGray & Grayscale & 0.4494 & no & yes \\
dFV\_RX & Kinematics & 0.4387 & no & yes \\
V & Kinematics & 0.4209 & no & yes \\
FV\_R & Kinematics & 0.4102 & no & yes \\
FA & Kinematics & 0.3544 & no & yes \\
Taylor\_ECRX & Morphology & 0.3189 & no & yes \\
\hline
\end{tabular}

\end{table}

% Table S7
\begin{table}[htbp]
\centering
\caption{\textbf{Supplementary Table S7.} Complete classifier-wise validation statistics comparing the full 93-dimensional dynamic feature set and the stability-guided reduced subsets. Values are reported as mean $\pm$ s.d. across the train:test split conditions evaluated in Figure 5. $\Delta$ values indicate the performance difference between the stability-guided subset and the full dynamic feature set. Significance was evaluated using the paired Wilcoxon signed-rank test.}
\label{tab:S7}
\resizebox{\textwidth}{!}{\small
\begin{tabular}{lllcccccccc}
\hline
Task & Classifier & Feature set & Accuracy & Macro-F1 & $\Delta$Acc & $p_{\mathrm{Acc}}$ & Sig. & $\Delta$F1 & $p_{\mathrm{F1}}$ & Sig. \\
\hline
Cancer-versus-healthy & SVM & Full dynamic feature set & 0.924 ± 0.006 & 0.900 ± 0.008 &  &  &  &  &  &  \\
Cancer-versus-healthy & SVM & Stability-guided subset & 0.902 ± 0.005 & 0.871 ± 0.007 & -0.0220 & <0.0001 & *** & -0.0297 & <0.0001 & *** \\
Cancer-versus-healthy & RF & Full dynamic feature set & 0.892 ± 0.006 & 0.849 ± 0.010 &  &  &  &  &  &  \\
Cancer-versus-healthy & RF & Stability-guided subset & 0.889 ± 0.005 & 0.848 ± 0.008 & -0.0034 & 0.0057 & ** & -0.0012 & 0.5612 & ns \\
Cancer-versus-healthy & LR & Full dynamic feature set & 0.870 ± 0.003 & 0.829 ± 0.005 &  &  &  &  &  &  \\
Cancer-versus-healthy & LR & Stability-guided subset & 0.821 ± 0.003 & 0.753 ± 0.004 & -0.0493 & <0.0001 & *** & -0.0767 & <0.0001 & *** \\
Cancer-versus-healthy & KNN & Full dynamic feature set & 0.851 ± 0.006 & 0.797 ± 0.008 &  &  &  &  &  &  \\
Cancer-versus-healthy & KNN & Stability-guided subset & 0.876 ± 0.006 & 0.832 ± 0.009 & +0.0243 & <0.0001 & *** & +0.0353 & <0.0001 & *** \\
\hline
Cancer subtype & SVM & Full dynamic feature set & 0.974 ± 0.003 & 0.973 ± 0.003 &  &  &  &  &  &  \\
Cancer subtype & SVM & Stability-guided subset & 0.976 ± 0.002 & 0.976 ± 0.002 & +0.0025 & 0.0022 & ** & +0.0026 & 0.0028 & ** \\
Cancer subtype & RF & Full dynamic feature set & 0.937 ± 0.004 & 0.936 ± 0.004 &  &  &  &  &  &  \\
Cancer subtype & RF & Stability-guided subset & 0.933 ± 0.006 & 0.932 ± 0.006 & -0.0046 & <0.0001 & *** & -0.0048 & <0.0001 & *** \\
Cancer subtype & LR & Full dynamic feature set & 0.971 ± 0.003 & 0.970 ± 0.003 &  &  &  &  &  &  \\
Cancer subtype & LR & Stability-guided subset & 0.964 ± 0.001 & 0.963 ± 0.001 & -0.0066 & <0.0001 & *** & -0.0068 & <0.0001 & *** \\
Cancer subtype & KNN & Full dynamic feature set & 0.939 ± 0.004 & 0.938 ± 0.004 &  &  &  &  &  &  \\
Cancer subtype & KNN & Stability-guided subset & 0.947 ± 0.004 & 0.946 ± 0.004 & +0.0084 & <0.0001 & *** & +0.0081 & <0.0001 & *** \\
\hline
\end{tabular}

% Note: $\Delta$ values are calculated as stability-guided subset minus full dynamic feature set.
% Significance labels are from the Wilcoxon signed-rank tests used in Figure 5b,c.
}
\end{table}

% Table S8: already contains longtable, caption, and label
\clearpage
%\input{SI/Table_S8_cross_flow_structural_consistency_selected_features}

%\clearpage

\end{document}